\documentclass[sn-basic,iicol]{sn-jnl}
\usepackage{amsmath,amsfonts}
\usepackage{graphicx}%
\usepackage{multirow}%
\usepackage{amsmath,amssymb,amsfonts}%
\usepackage{amsthm}%
\usepackage{mathrsfs}%
\usepackage[title]{appendix}%
\usepackage{xcolor}%
\usepackage{textcomp}%
\usepackage{manyfoot}%
\usepackage{booktabs}%
\usepackage{algorithm}%
\usepackage{algorithmicx}%
\usepackage{algpseudocode}%
\usepackage{color}
\usepackage{listings}%
\usepackage{array}    %
\usepackage{graphicx}
\usepackage{subcaption}
\usepackage{longtable} %
\usepackage{ragged2e}
\usepackage{pifont}       
\usepackage{bbding}       

\theoremstyle{thmstyleone}%
\theoremstyle{thmstyletwo}%

\theoremstyle{thmstylethree}%

\begin{document}

\makeatletter
\newcommand{\rmnum}[1]{\romannumeral #1}
\newcommand{\Rmnum}[1]{\expandafter\@slowromancap\romannumeral #1@}
\makeatother

\newcommand{\reviseMark}[1]{\textcolor{black}{#1}}

\title[Article Title]{Image-Based Virtual Try-On: A Survey}


\author[1]{\fnm{Dan} \sur{Song}}\email{dan.song@tju.edu.cn}
\author[1]{\fnm{Xuanpu} \sur{Zhang}}\email{misfit\_echo@tju.edu.cn}
\author[1]{\fnm{Juan} \sur{Zhou}}\email{zhoujuan@tju.edu.cn}
\author[1]{\fnm{Weizhi} \sur{Nie}}\email{weizhinie@tju.edu.cn}
\author[2]{\fnm{Ruofeng} \sur{Tong}}\email{trf@zju.edu.cn}
\author[3]{\fnm{Mohan} \sur{Kankanhalli}}\email{mohan@comp.nus.edu.sg}
\author*[1]{\fnm{An-An} \sur{Liu}}\email{anan0422@gmail.com}

\affil*[1]{\orgdiv{The School of Electrical and Information Engineering}, \orgname{Tianjin University}, \orgaddress{\postcode{300072}, \country{China}}}

\affil[2]{\orgdiv{The College of Computer Science and Technology}, \orgname{Zhejiang University}, \orgaddress{\postcode{310007}, \country{China}}}

\affil[3]{\orgdiv{The School of Computing}, \orgname{National University of Singapore}, \orgaddress{ \postcode{117543}, \country{Singapore}}}


\abstract{Image-based virtual try-on aims to synthesize a naturally dressed person image with a clothing image, which revolutionizes online shopping and inspires related topics within image generation, showing both research significance and commercial potential. 
However, there is a gap between current research progress and commercial applications and an absence of comprehensive overview of this field to accelerate the development.
In this survey, we provide a comprehensive analysis of the state-of-the-art techniques and methodologies in aspects of pipeline architecture, person representation and key modules such as try-on indication, clothing warping and try-on stage. 
\reviseMark{We additionally apply CLIP to assess the semantic alignment of try-on results}, and evaluate representative methods with uniformly implemented evaluation metrics on the same dataset.
In addition to quantitative and qualitative evaluation of current open-source methods, 
unresolved issues are highlighted and future research directions are prospected to identify key trends and inspire further exploration. 
The uniformly implemented evaluation metrics, dataset and collected methods will be made public available at https://github.com/little-misfit/Survey-Of-Virtual-Try-On.}




\maketitle

\section{Introduction}\label{sec1}
Image-based virtual try-on is a popular research topic in the field of AI-generated content (AIGC), specifically in the domain of conditional person image generation. It enables editing, replacement, and design of clothing image content, making it highly applicable in various domains such as e-commerce platforms and short video platforms. In particular, online shoppers can benefit from virtual try-on by obtaining try-on effect images of clothing, thereby enhancing their shopping experience and increasing the likelihood of successful transactions. In addition, AI Fashion has also emerged on short video platforms, where users can edit the clothes worn by characters in images or videos according to their own creativity. This allows users to explore their sense of fashion and produce a wide range of engaging images and videos.
\begin{figure}[t]
\centering
\includegraphics[width=3in]{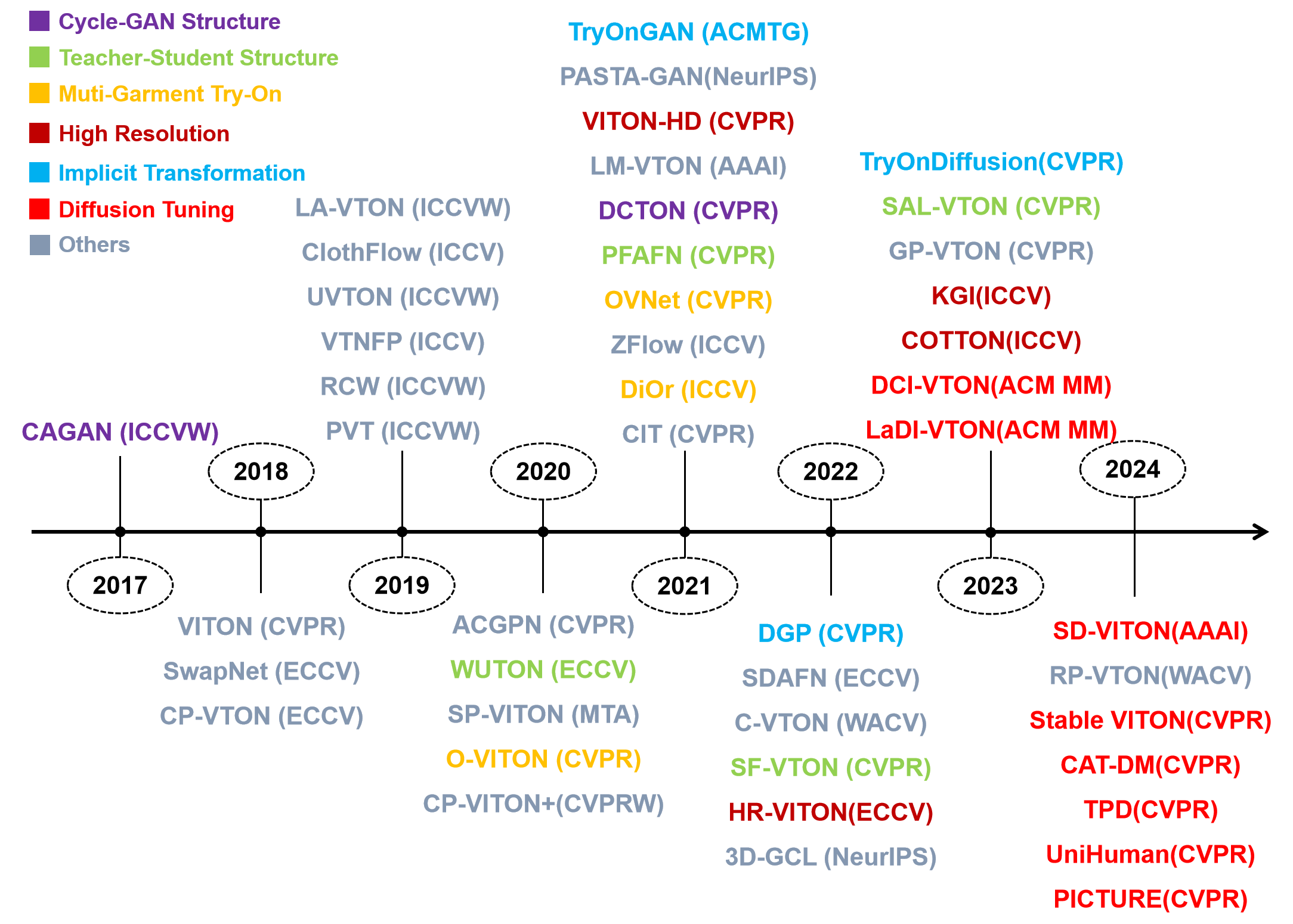}
\caption{\reviseMark{A concise timeline of image-based virtual try-on milestones. Different colors indicate the main characteristic of method. Please refer to Table \ref{big_table} for detailed comparisons. }}
\label{Time_line}
\end{figure}

The concept of virtual try-on was proposed as early as 2001  \citep{zuizaodeshiyi}, which uses a pre-calculated generic database to produce personally sized bodies and animate garments on a web application. 
Virtual try-on methods can be divided into three categories: physical-based simulation, real acquisition and image generation. Based on the cloth simulation techniques  \citep{physical_fushifangzhen_1,physical_fushifangzhen_2,physical_fushifangzhen_3} as the core part, physical-based try-on simulation also involves human body reconstruction  \citep{physical_body_1,physical_body_2,physical_body_3,physical_body_4,physical_body_5,physical_body_6,physical_body_7,physical_body_8,physical_body_9,physical_body_10} and cloth measurement  \citep{physical_fushiceliang_1,physical_fushiceliang_2,physical_fushiceliang_3}. This kind of methods have the advantage in pose controllable and $360^\circ$ display, but face plenty of difficulties in accuracy, efficiency and automation. Some brands, e.g., UNIQLO and GAP, employ this way as a virtual try-on solution. Real acquisition method usually captures and stores the appearance of apparel on a shape controllable robot for later displays, which shows high fidelity. Only a few companies, such as Fits.me, have adopted this approach, which requires massive labor for real acquisition. With the development of image generation techniques, image-based virtual try-on becomes appealing with benefits in high efficiency and low cost. However, the artifacts in generated images hinder its application to practical scenarios. Specifically, this paper focuses on image-based virtual try-on methods with a thorough review in terms of methodology design and experimental evaluation, and further reveals unresolved issues leading to prospective future research directions.

Image-based virtual try-on can be regarded as one kind of conditional person image generation, which have undergone rapid development since 2017 (Fig. \ref{Time_line}). Given a clothed person image and an in-shop clothing image, image-based virtual try-on aims to synthesize a naturally dressed body image. For this task, three main difficulties need to be overcome: 1) Obtaining high-quality supervised training data: It is almost impossible to acquire pairs of photographs where the same person is trying on two different garments in the same pose. 2) Achieving natural and realistic bending and shading of clothing in appropriate areas of the person body: It is challenging to ensure that the clothing adapts seamlessly and naturally to the contours of the body and appears realistic in terms of lighting and shadows. 3) Generating realistic try-on images: It is necessary to maintain consistency in non-clothing areas such as removing the original clothing residual uncovered by the new clothing and keep the person identity clear.

To overcome the above difficulties, tremendous efforts have been made and Fig. \ref{Time_line} show some representative methods on a timeline. In 2017, CAGAN \citep{CAGAN_jiandanGAN} gave the first try by employing CycleGAN \citep{CycleGAN} to overcome the lack of training triplet data, i.e., (original person image, in-shop clothing image, try-on image),  but the generation quality is far from satisfactory.
Subsequently, VITON \citep{VITON} creatively proposed clothing-agnostic person representation by human parsing to make up the lack of supervised training data. They constructed the basic network framework of ``Try-On Indication + Cloth Warping + Try-on", laying the foundation for further improvement on generation quality in subsequent works  \citep{CP-VTON+,CP-VTON,LM-VITON,sp-VITON,ACGPN,RCW_CP,PVT_CP,LA-VTON,vtnfp,HR-VTON}. 
Recently, with the advancement of computational capabilities, high-resolution virtual try-on tasks has become possible, e.g., VITON-HD  \citep{VITON-HD} and HR-VTON  \citep{HR-VTON}. 
Inspired by StyleGAN  \citep{stylegan,stylegan1,stylegan2} and Diffusion model  \citep{DDPM,IDDPM,DDIM,InterGen} in the field of image generation, single-stage networks  \citep{Tryongan_yuduiqi,DGP_yuduiqi_ESF} as well as  Diffusion architecture  \citep{tryondiffusion, Stable-VITON, CAT-DM} emerged for this task.

\reviseMark{
Apart from continuous quality improvements, some new goals are desired. 
In order to remove the heavy reliance on human parsing, several teacher-student networks were designed \citep{WUTON,PFAFN,style-flow} to achieve parser-free try-on at inference time. 
Paired training requires tremendous paired data and does not fit well in the unpaired try-on scenario. Previous efforts such as interpolation in latent space \citep{Tryongan_yuduiqi}, uv mapping via dense pose \citep{street} and disentanglement of garment style and spatial features \citep{PASTA-GAN,PASTA-GAN-plus} have been made, to facilitate unpaired image-based virtual try-on. 
Trying on multiple garments, that is, mix-match, has practical values but is challenging in dealing with the interaction between garments (e.g., tuck or untuck). Generating semantic layout \citep{o-VTON, OVNet} or giving the dressing order \citep{DiOr,PASTA-GAN-plus} for different garments are two dominant solutions. 
Controllable try-on in regards of clothing sizes \citep{size_dose_matter} and types \citep{SAL-VTON, RP-VTON} are achieved via associated skeleton or landmarks. 
With the development of mask inpainting in the image generation area, recently in-the-wild try-on becomes popular \citep{UniHuman,street,PICTURE,Stable-VITON}, where the background is more natural and complex.
}

In spite of the rapidly emerging works, there lacks a systematic survey to summarize image-based virtual try-on methods in datasets, method design and experimental evaluation. Two previous reviews  \citep{zongshu1,zongshu2} only introduced several representative methods, but did not perform comprehensive comparison and unified evaluation. To the best of our knowledge, this is the first systematic image-based virtual try-on review with unified evaluation, which also has the following unique characteristics:
\begin{itemize}
    \item \textbf{In-depth analysis.}  We comprehensively review existing image-based virtual try-on methods from the perspectives of pipeline structures, human representations, clothing warping strategies, architectures of try-on indication and image synthesis and corresponding loss functions.
    \item \textbf{Evaluation: unified evaluation, new criterion, and user study.} We evaluate open-source works with the same dataset, and also perform a user study on visual results with \reviseMark{139} volunteers. 
    \reviseMark{Particularly, we separately evaluate try-on and non-try-on parts and additionally apply CLIP \citep{clip} to measure the semantic similarity. }
    Relevant data and codes will be publicly available at https://github.com/little-misfit/Survey-Of-Virtual-Try-On.
    \item \textbf{Open challenges and future directions.} According to the experimental results, we reveal a number of unresolved issues and draw important future research directions. 
    We hope this review could spur the development of novel ideas towards image-based virtual try-on and its applications in industry. 
\end{itemize}

The rest of this survey is organized as follows. Sec. \ref{Technical review and discussion} firstly gives the problem definition, and then comprehensively review the literature from multiple perspectives. Sec. \ref{Datasets and Evaluation Criteria} introduces datasets and evaluation criterion. Experimental results and analyses are presented in Sec. \ref{sec:experiments}. Finally, we reveal unresolved issues in Sec. \ref{sec:unresolved} and prospect future directions in Sec. \ref{sec:future}. 



\begin{table*}[htbp]
\begin{center}
\caption{Key characteristics of representative image-based virtual try-On methods.  Please refer the number of pipeline to Fig. \ref{totalmodel} and the number of person representation is shown in Fig. \ref{fig:person_representation}. The three loss items are the respective losses for modules of Try-On Indication, Cloth Warping and Try-on. 
}
\label{big_table}
\resizebox{1\textwidth}{!}{
\setlength\tabcolsep{2.5pt}
\renewcommand\arraystretch{1}
\begin{tabular}{c c c c c c c c c c c c c c c c c c c c c}
\toprule
\multicolumn{1}{c}{\textbf{Year}} & \multicolumn{1}{c}{\textbf{Model}} & \multicolumn{1}{c}{\textbf{Source}} & \multicolumn{1}{c}{\textbf{Pipeline}} & \multicolumn{1}{c}{\textbf{Person}} & \multicolumn{1}{c}{\textbf{Try-On}} & \multicolumn{4}{@{}c@{}}{\textbf{Loss}} & \multicolumn{1}{c}{\textbf{Cloth}} & \multicolumn{4}{@{}c@{}}{\textbf{Loss}} & \multicolumn{1}{c}{\textbf{Try-On}} & \multicolumn{4}{@{}c@{}}{\textbf{Loss}} & \multicolumn{1}{c}{\textbf{Train/Test}} \\ 
 & & & & \textbf{Rep.} & \textbf{Indication} & \textbf{Per} & \textbf{L1} &  \textbf{Adv} & \textbf{Ent} & \textbf{Warping} & \textbf{Per} & \textbf{L1} &  \textbf{Adv} & \textbf{Reg} & & \textbf{Per} & \textbf{L1} &  \textbf{Adv} & \textbf{Oths} & \textbf{Dataset}\\
\midrule

\multicolumn{1}{c}{\multirow{1}{*}{2017}} & \begin{tabular}[c]{@{}c@{}}CAGAN\\  \citep{CAGAN_jiandanGAN} \end{tabular} & ICCVW & {\Rmnum{1}} & \begin{tabular}[c]{@{}c@{}}1\end{tabular} & - & \multicolumn{4}{c}{-} & - & \multicolumn{4}{c}{-} & \begin{tabular}[c]{@{}c@{}} Cycle-GAN\\Mask Combine\end{tabular} & &\checkmark &\checkmark &\checkmark & Zalando SE \\ 

\multicolumn{1}{c}{\multirow{3}{*}{2018}} 
&\begin{tabular}[c]{@{}c@{}}VITON\\  \citep{VITON} \end{tabular} & CVPR & {\Rmnum{5}} & \begin{tabular}[c]{@{}c@{}}2,5,6 \end{tabular} & U-Net & \checkmark &\checkmark & & & TPS & \multicolumn{4}{c}{-} & Mask Combine & \checkmark &\checkmark & & \checkmark & VITON\\ 

&\begin{tabular}[c]{@{}c@{}}CP-VTON\\  \citep{CP-VTON} \end{tabular}& ECCV & {\Rmnum{6}} & \begin{tabular}[c]{@{}c@{}}2,5,6 \end{tabular} & U-Net & \multicolumn{4}{c}{-} & TPS & & \checkmark & & & Mask Combine & \checkmark &\checkmark & & & VITON\\ 

& \begin{tabular}[c]{@{}c@{}}SwapNet\\  \citep{swapnet} \end{tabular}& ECCV & {\Rmnum{2}} & \begin{tabular}[c]{@{}c@{}}2,10 \end{tabular} & \begin{tabular}[c]{@{}c@{}}Dualpath-\\ U-Net \end{tabular} & \checkmark & \checkmark & & & - & \multicolumn{4}{c}{-} & U-Net & \checkmark & \checkmark & \checkmark & & \begin{tabular}[c]{@{}c@{}}DeepFashion\\ / VITON\end{tabular}\\   



\multirow{4}*{2019} 
&\begin{tabular}[c]{@{}c@{}}LA-VTON\\  \citep{LA-VTON} \end{tabular}& CVPR & {\Rmnum{6}} & \begin{tabular}[c]{@{}c@{}}2,5,7 \end{tabular} & U-Net & &\checkmark & & & TPS &  &\checkmark &\checkmark &\checkmark & \begin{tabular}[c]{@{}c@{}}Mask Combine\\Refine Net\end{tabular} &\checkmark &\checkmark &\checkmark & & VITON\\ 

&\begin{tabular}[c]{@{}c@{}}VTNFP\\  \citep{vtnfp} \end{tabular} & ICCV & {\Rmnum{6}} & \begin{tabular}[c]{@{}c@{}}2,4,5,11 \end{tabular} & CNN & & &\checkmark &  & TPS & &\checkmark &\checkmark & & \begin{tabular}[c]{@{}c@{}}Attention U-Net\\ Mask Combine \end{tabular} &\checkmark &\checkmark & & & VITON\\  

&\begin{tabular}[c]{@{}c@{}}Clothflow\\  \citep{clothflow_flownet} \end{tabular} & ICCV & {\Rmnum{5}} & \begin{tabular}[c]{@{}c@{}}1,5,10 \end{tabular} & U-Net & & & &\checkmark& FlowNet &\checkmark &\checkmark & &\checkmark & \begin{tabular}[c]{@{}c@{}}U-Net\end{tabular} &\checkmark &  & &\checkmark & \begin{tabular}[c]{@{}c@{}}DeepFashion\\ / VITON\end{tabular}\\ 

&\begin{tabular}[c]{@{}c@{}}UVTON\\  \citep{u-vton} \end{tabular} & ICCVW & {\Rmnum{3}} & \begin{tabular}[c]{@{}c@{}}2,9 \end{tabular} & Multi-GAN &\checkmark &\checkmark &\checkmark & & - & \multicolumn{4}{c}{-} & U-Net &\checkmark &\checkmark &\checkmark & & VITON\\  

\multicolumn{1}{c}{\multirow{5}{*}{2020}}
&\begin{tabular}[c]{@{}c@{}}SP-VITON\\  \citep{sp-VITON} \end{tabular} & MTAP & {\Rmnum{5}} & \begin{tabular}[c]{@{}c@{}}2,5,8 \end{tabular} & U-Net & \checkmark & \checkmark & & & TPS & \multicolumn{4}{c}{-} & \begin{tabular}[c]{@{}c@{}}U-Net\\ Mask Combine\end{tabular}& \checkmark &\checkmark & &\checkmark& VITON\\  

&\begin{tabular}[c]{@{}c@{}}ACGPN\\  \citep{ACGPN} \end{tabular} & CVPR & {\Rmnum{5}} & \begin{tabular}[c]{@{}c@{}}2,4,11 \end{tabular} & Multi-GAN &  & & \checkmark &\checkmark & TPS &  &\checkmark & &\checkmark & \begin{tabular}[c]{@{}c@{}}Body Part-\\Composition GAN\end{tabular}&  & & \checkmark & & VITON\\  

&\begin{tabular}[c]{@{}c@{}}O-VITON\\  \citep{o-VTON} \end{tabular}& CVPR & {\Rmnum{5}} & \begin{tabular}[c]{@{}c@{}}9,10 \end{tabular} & Auto-Encoder &  & & \checkmark &\checkmark& - & \multicolumn{4}{c}{-} & \begin{tabular}[c]{@{}c@{}} Broadcast  Decoder \end{tabular}& \checkmark & \checkmark & & & VITON\\ 

&\begin{tabular}[c]{@{}c@{}}CP-VTON+\\  \citep{CP-VTON+} \end{tabular}& CVPRW & {\Rmnum{6}} & \begin{tabular}[c]{@{}c@{}}2,5,7 \end{tabular} & U-Net &  & \checkmark & & & TPS &   & \checkmark & &\checkmark & \begin{tabular}[c]{@{}c@{}}U-Net\\ Mask Combine \end{tabular}& \checkmark & \checkmark & & & VITON\\  

&\begin{tabular}[c]{@{}c@{}}WUTON\\  \citep{WUTON} \end{tabular} & ECCV & {\Rmnum{4}} & \begin{tabular}[c]{@{}c@{}}3 \end{tabular} & - & \multicolumn{4}{c}{-}& TPS &  &\checkmark & &  & \begin{tabular}[c]{@{}c@{}}Siamese U-Net \end{tabular}& \checkmark &\checkmark &\checkmark & & VITON\\  

\multicolumn{1}{c}{\multirow{10}{*}{2021}}
&\begin{tabular}[c]{@{}c@{}}TryOnGAN\\  \citep{Tryongan_yuduiqi} \end{tabular} & TOG & {\Rmnum{2}} & \begin{tabular}[c]{@{}c@{}}1 \end{tabular} & - & \multicolumn{4}{c}{-} & \begin{tabular}[c]{@{}c@{}}Implicit\\Transform \end{tabular} & \multicolumn{4}{c}{-} & \begin{tabular}[c]{@{}c@{}}StyleGAN2 \end{tabular}& \checkmark & & &\checkmark & Self-Made\\ 

&\begin{tabular}[c]{@{}c@{}}LM-VTON\\  \citep{LM-VITON} \end{tabular} & AAAI & {\Rmnum{6}} & \begin{tabular}[c]{@{}c@{}}2,5,11 \end{tabular} & U-Net &  & & \checkmark &\checkmark& TPS & \checkmark &\checkmark & &\checkmark & \begin{tabular}[c]{@{}c@{}}U-Net \end{tabular}& \checkmark &\checkmark &\checkmark & & \begin{tabular}[c]{@{}c@{}}VITON\\ MPV\end{tabular}\\  

&\begin{tabular}[c]{@{}c@{}}OVNet\\  \citep{OVNet} \end{tabular}  & CVPR & {\Rmnum{5}} & \begin{tabular}[c]{@{}c@{}}5,12 \end{tabular} & U-Net &  & & \checkmark &\checkmark & Multi-STN &   & \checkmark & & \checkmark  & \begin{tabular}[c]{@{}c@{}}U-Net \end{tabular}& \checkmark &\checkmark &\checkmark & & \begin{tabular}[c]{@{}c@{}}Self Made\\ / VITON\end{tabular}\\  

&\begin{tabular}[c]{@{}c@{}}CIT\\  \citep{clothflow}  \end{tabular}& CVPR & {\Rmnum{6}} & \begin{tabular}[c]{@{}c@{}}4,5,7 \end{tabular} & \begin{tabular}[c]{@{}c@{}}Transformer-\\ U-Net \end{tabular} &  & \checkmark & & & \begin{tabular}[c]{@{}c@{}}Transformer-\\ TPS \end{tabular} &   & \checkmark & &\checkmark & \begin{tabular}[c]{@{}c@{}}U-Net\\ Mask Combine \end{tabular}& \checkmark &\checkmark & & & VITON\\  

&\begin{tabular}[c]{@{}c@{}}DCTON\\  \citep{DCTON} \end{tabular} & CVPR & {\Rmnum{4}} & \begin{tabular}[c]{@{}c@{}}1,8 \end{tabular} & - & \multicolumn{4}{c}{-} & TPS &   & \checkmark & &\checkmark & \begin{tabular}[c]{@{}c@{}}CNN \end{tabular}& \checkmark &\checkmark &\checkmark &\checkmark & VITON\\  

&\begin{tabular}[c]{@{}c@{}}PFAFN\\  \citep{PFAFN} \end{tabular} & CVPR & {\Rmnum{4}} & \begin{tabular}[c]{@{}c@{}}5,8,12 \end{tabular} & - & \multicolumn{4}{c}{-} & FlowNet &   & & &\checkmark & \begin{tabular}[c]{@{}c@{}}U-Net\end{tabular}& \checkmark &\checkmark & & & \begin{tabular}[c]{@{}c@{}}MPV\\ VITON\end{tabular}\\  

&\begin{tabular}[c]{@{}c@{}}VITON-HD\\  \citep{VITON-HD} \end{tabular} & CVPR & {\Rmnum{5}} & \begin{tabular}[c]{@{}c@{}}3,5,12 \end{tabular} & U-Net &  & & \checkmark &\checkmark & TPS &   & & &\checkmark & \begin{tabular}[c]{@{}c@{}}ALIAS Generator \end{tabular}& \checkmark & &\checkmark &\checkmark & VITON-HD\\  

&\begin{tabular}[c]{@{}c@{}}DiOr\\  \citep{DiOr} \end{tabular}  & ICCV & {\Rmnum{4}} & \begin{tabular}[c]{@{}c@{}}1,5 \end{tabular} & - & \multicolumn{4}{c}{-} & FlowNet & \multicolumn{4}{c}{-} & \begin{tabular}[c]{@{}c@{}}Broadcast  Decoder \end{tabular}& \checkmark &\checkmark &\checkmark &\checkmark & DeepFashion\\  

&\begin{tabular}[c]{@{}c@{}}ZFlow\\  \citep{zflow} \end{tabular} & ICCV & {\Rmnum{7}} & \begin{tabular}[c]{@{}c@{}}4,5,7,8,9 \end{tabular} & U-Net &   & & &\checkmark & FlowNet & \checkmark &\checkmark & &\checkmark & \begin{tabular}[c]{@{}c@{}}U-Net\\ Mask Combine \end{tabular}& \checkmark &\checkmark & &\checkmark & VITON\\  

&\begin{tabular}[c]{@{}c@{}}\reviseMark{PASTA-GAN}\\  \citep{PASTA-GAN} \end{tabular} & \reviseMark{NeurIPS} & {\reviseMark{\Rmnum{6}}} & \begin{tabular}[c]{@{}c@{}}\reviseMark{1,2,5} \end{tabular} & \reviseMark{GAN} &  \reviseMark{\checkmark} & \reviseMark{\checkmark} & \reviseMark{\checkmark} & & \reviseMark{Multi-STN} & \multicolumn{4}{c}{\reviseMark{-}} & \begin{tabular}[c]{@{}c@{}}\reviseMark{GAN} \end{tabular}& \reviseMark{\checkmark} & \reviseMark{\checkmark} & \reviseMark{\checkmark} & & \begin{tabular}[c]{@{}c@{}}\reviseMark{DeepFashion}\\ \reviseMark{MPV}\\ \reviseMark{VPT}\end{tabular}\\

\multicolumn{1}{c}{\multirow{7}{*}{2022}}
&\begin{tabular}[c]{@{}c@{}}C-VTON\\  \citep{C-VTON} \end{tabular} & WACV & {\Rmnum{4}} & \begin{tabular}[c]{@{}c@{}}3,8 \end{tabular} & - & \multicolumn{4}{c}{-} & TPS & \checkmark &\checkmark & & & \begin{tabular}[c]{@{}c@{}}U-Net \end{tabular}& \checkmark & &\checkmark & & \begin{tabular}[c]{@{}c@{}}MPV\\ VITON\end{tabular}\\  

&\begin{tabular}[c]{@{}c@{}}Flow-Style-VTON \\  \citep{style-flow} \end{tabular} & CVPR & {\Rmnum{4}} & \begin{tabular}[c]{@{}c@{}}5,8,12 \end{tabular} & - & \multicolumn{4}{c}{-} & FlowNet & \multicolumn{4}{c}{-} & \begin{tabular}[c]{@{}c@{}}U-Net \end{tabular}&   & \checkmark & &\checkmark & VITON\\  

&\begin{tabular}[c]{@{}c@{}}DGP\\  \citep{DGP_yuduiqi_ESF} \end{tabular} & CVPR & {\Rmnum{2}} & \begin{tabular}[c]{@{}c@{}}13 \end{tabular} & - & \multicolumn{4}{c}{-} & \begin{tabular}[c]{@{}c@{}}Implicit\\Transform\end{tabular} & \multicolumn{4}{c}{-} & \begin{tabular}[c]{@{}c@{}}Decoder \end{tabular}& \checkmark &\checkmark &\checkmark &\checkmark & \begin{tabular}[c]{@{}c@{}}ESF\\ / CMI MPV\end{tabular}\\  

&\begin{tabular}[c]{@{}c@{}}RT-VTON\\  \citep{RT-VTON} \end{tabular} & CVPR & {\Rmnum{5}} & \begin{tabular}[c]{@{}c@{}}1,5,10 \end{tabular} & \begin{tabular}[c]{@{}c@{}} CNN \end{tabular} &  & & & & TPS & \checkmark &\checkmark & &\checkmark & \begin{tabular}[c]{@{}c@{}} U-Net \end{tabular}& \checkmark & \checkmark &\checkmark & & VITON\\ 

&\begin{tabular}[c]{@{}c@{}}HR-VITON\\  \citep{HR-VTON} \end{tabular} & ECCV & {\Rmnum{7}} & \begin{tabular}[c]{@{}c@{}}3,8,12 \end{tabular} & \begin{tabular}[c]{@{}c@{}}Dualpath-\\ U-Net \end{tabular} &  & & &\checkmark & FlowNet & \checkmark &\checkmark &\checkmark &\checkmark & \begin{tabular}[c]{@{}c@{}} Mask Combine\\Decoder \end{tabular}& \checkmark & &\checkmark &\checkmark & VITON-HD\\  

&\begin{tabular}[c]{@{}c@{}}SDAFN\\  \citep{SDFN} \end{tabular} & ECCV & {\Rmnum{4}} & \begin{tabular}[c]{@{}c@{}}3,4 \end{tabular} & - & \multicolumn{4}{c}{-} & \begin{tabular}[c]{@{}c@{}}Multi-\\ FlowNet\end{tabular} & \multicolumn{4}{c}{-} & \begin{tabular}[c]{@{}c@{}}Decoder \end{tabular}& \checkmark &\checkmark & &\checkmark & \begin{tabular}[c]{@{}c@{}}MPV\\ VITON\end{tabular}\\ 

 & \begin{tabular}[c]{@{}c@{}}3D-GCL\\  \citep{3D-GCL} \end{tabular} & NeurIPS & {\Rmnum{4}} & \begin{tabular}[c]{@{}c@{}}9\end{tabular} & - & 
\multicolumn{4}{c}{-} & FlowNet & \checkmark &\checkmark & &\checkmark & \begin{tabular}[c]{@{}c@{}}StyleGAN2\end{tabular} & \checkmark &\checkmark &\checkmark & & \begin{tabular}[c]{@{}c@{}}MPV\\ DeepFashion\end{tabular} \\ 

\multicolumn{1}{c}{\multirow{7}{*}{2023}}
& \begin{tabular}[c]{@{}c@{}}POVNet\\  \citep{pami-POVNet} \end{tabular}  & TPAMI & {\Rmnum{5}} & \begin{tabular}[c]{@{}c@{}}2,5,12\end{tabular}& U-Net & & & \checkmark& \checkmark&  Multi-STN& &\checkmark & & \checkmark &U-Net  &\checkmark  &\checkmark &\checkmark & &\begin{tabular}[c]{@{}c@{}}Dress-Code\\VITON-HD\end{tabular}\\ 

&\begin{tabular}[c]{@{}c@{}}GP-VTON\\  \citep{GP-VTON} \end{tabular} & CVPR & {\Rmnum{7}} & \begin{tabular}[c]{@{}c@{}}5,8,12 \end{tabular} & \begin{tabular}[c]{@{}c@{}}Dualpath-\\ U-Net \end{tabular} & & & \checkmark & \checkmark & \begin{tabular}[c]{@{}c@{}}Multi-\\ FlowNet\end{tabular} & \checkmark &\checkmark & &\checkmark & \begin{tabular}[c]{@{}c@{}} Mask Combine\\+U-Net \end{tabular}& \checkmark &\checkmark &\checkmark & & \begin{tabular}[c]{@{}c@{}}VITON-HD\\Dress Code\end{tabular}\\  

&\begin{tabular}[c]{@{}c@{}}SAL-VTON\\  \citep{SAL-VTON} \end{tabular} & CVPR & {\Rmnum{4}} & \begin{tabular}[c]{@{}c@{}}1,13 \end{tabular} & \begin{tabular}[c]{@{}c@{}}-\end{tabular} & \multicolumn{4}{c}{-} & \begin{tabular}[c]{@{}c@{}}Multi-\\ FlowNet\end{tabular} & \checkmark &\checkmark & &\checkmark & \begin{tabular}[c]{@{}c@{}} U-Net \end{tabular}& \checkmark & &\checkmark &\checkmark & \begin{tabular}[c]{@{}c@{}}VITON\\ VITON-HD\end{tabular}\\

&\begin{tabular}[c]{@{}c@{}}TryOnDiffusion\\  \citep{tryondiffusion} \end{tabular} & CVPR & {\Rmnum{1}} & \begin{tabular}[c]{@{}c@{}}4 \end{tabular} & - & \multicolumn{4}{c}{-} & - & \multicolumn{4}{c}{-} & Diffusion & & & & \checkmark & \begin{tabular}[c]{@{}c@{}}MPV\\VITON\end{tabular}\\

& \begin{tabular}[c]{@{}c@{}}KGI\\  \citep{KGI} \end{tabular}  & ICCV &{\Rmnum{1}}& \begin{tabular}[c]{@{}c@{}}3,5,12,13\end{tabular} & Auto-Encoder & & & & \checkmark & TPS & & \checkmark & & & Diffusion &  & & &\checkmark &VITON-HD  \\

& \begin{tabular}[c]{@{}c@{}}COTTON\\  \citep{size_dose_matter} \end{tabular}  & ICCV & {\Rmnum{1}} & \begin{tabular}[c]{@{}c@{}}4,5,10 \end{tabular}& U-Net &  & & & \checkmark & STN &  & \checkmark & & & U-Net & \checkmark & \checkmark & \checkmark &   & \begin{tabular}[c]{@{}c@{}}COTTON\\VITON-HD\end{tabular} \\ 

& \begin{tabular}[c]{@{}c@{}}Ladi-VTON\\  \citep{LaDI-VTON} \end{tabular}  & ACM MM & {\Rmnum{1}} & \begin{tabular}[c]{@{}c@{}}3,5 \end{tabular}&- &\multicolumn{4}{c}{-} & - & \multicolumn{4}{c}{-} & Diffusion & \checkmark  & \checkmark & & \checkmark& \begin{tabular}[c]{@{}c@{}}Dress-Code\\ VITON-HD \end{tabular} \\ 

& \begin{tabular}[c]{@{}c@{}}DCI-VTON\\  \citep{DCI-VTON} \end{tabular}  & ACM MM & {\Rmnum{4}} & \begin{tabular}[c]{@{}c@{}}3, 8,12 \end{tabular}&- &\multicolumn{4}{c}{-} & FlowNet &\checkmark &\checkmark & & & Diffusion &\checkmark  & & & \checkmark& \begin{tabular}[c]{@{}c@{}}Dress-Code\\ VITON-HD \end{tabular} \\ 

\multicolumn{1}{c}{\multirow{6}{*}{2024}}  & \begin{tabular}[c]{@{}c@{}}SD-VITON\\  \citep{SD-VITON} \end{tabular} & AAAI & {\Rmnum{7}} & \begin{tabular}[c]{@{}c@{}}3,8,12\end{tabular}  &\begin{tabular}[c]{@{}c@{}}Dualpath-\\ U-Net \end{tabular}  & & & & \checkmark& FlowNet & \checkmark& \checkmark & \checkmark& \checkmark & \begin{tabular}[c]{@{}c@{}} Mask Combine\\Decoder \end{tabular}& \checkmark & &\checkmark &\checkmark & VITON-HD\\  

&\begin{tabular}[c]{@{}c@{}}Stable-VITON\\  \citep{Stable-VITON} \end{tabular} & CVPR & {\Rmnum{1}} & \begin{tabular}[c]{@{}c@{}}3,8 \end{tabular} &-& \multicolumn{4}{c}{-} &-  &  \multicolumn{4}{c}{-}& \begin{tabular}[c]{@{}c@{}}Diffusion\end{tabular}&  & & & \checkmark&\begin{tabular}[c]{@{}c@{}}Dress-Code\\VITON-HD\end{tabular}\\ 

 & \begin{tabular}[c]{@{}c@{}}CAT-DM\\  \citep{CAT-DM} \end{tabular} & CVPR & {\Rmnum{1}} & \begin{tabular}[c]{@{}c@{}}3,8\end{tabular}  &- & \multicolumn{4}{c}{-}&-  & \multicolumn{4}{c}{-} &Diffusion &  & & &\checkmark &\begin{tabular}[c]{@{}c@{}}Dress-Code\\VITON-HD\end{tabular}\\  

 & \begin{tabular}[c]{@{}c@{}}\reviseMark{TPD}\\  \citep{TPD} \end{tabular} & \reviseMark{CVPR} & {\reviseMark{\Rmnum{3}}} & \begin{tabular}[c]{@{}c@{}}\reviseMark{3,5,8}\end{tabular}  &\reviseMark{Diffusion} & \multicolumn{4}{c}{\reviseMark{-}} & \reviseMark{-} & \multicolumn{4}{c}{\reviseMark{-}} & \reviseMark{Diffusion} &  & & &\reviseMark{\checkmark} & \begin{tabular}[c]{@{}c@{}}\reviseMark{VITON}\\ \reviseMark{VITON-HD}\end{tabular}\\  

 & \begin{tabular}[c]{@{}c@{}}\reviseMark{UniHuman}\\  \citep{UniHuman} \end{tabular} & \reviseMark{CVPR} & {\reviseMark{\Rmnum{1}}} & \begin{tabular}[c]{@{}c@{}}\reviseMark{1,5,9,13}\end{tabular}  &\reviseMark{-} & \multicolumn{4}{c}{\reviseMark{-}}& \reviseMark{-}  & \multicolumn{4}{c}{\reviseMark{-}} &\reviseMark{Diffusion} &  &\reviseMark{\checkmark} & &\reviseMark{\checkmark} &\begin{tabular}[c]{@{}c@{}}\reviseMark{LH-400K}\\ \reviseMark{WPose}\end{tabular}\\  
 
 & \begin{tabular}[c]{@{}c@{}}\reviseMark{PICTURE}\\  \citep{PICTURE} \end{tabular} & \reviseMark{CVPR} & {\reviseMark{\Rmnum{3}}} & \begin{tabular}[c]{@{}c@{}}\reviseMark{4,8,12}\end{tabular}  &\reviseMark{Diffusion} & \multicolumn{4}{c}{\reviseMark{-}} & \reviseMark{-} & \multicolumn{4}{c}{\reviseMark{-}} & \reviseMark{Diffusion} &  & & &\reviseMark{\checkmark} & \begin{tabular}[c]{@{}c@{}}\reviseMark{WUTON}\\ \reviseMark{DeepFashionM}\\ \reviseMark{SHHQ}\\ \reviseMark{VITON-HD}\end{tabular}\\   


\bottomrule
\end{tabular}
}
\end{center}
\end{table*}

\begin{figure*}[!t]
\centering
\includegraphics[width=6.2in]{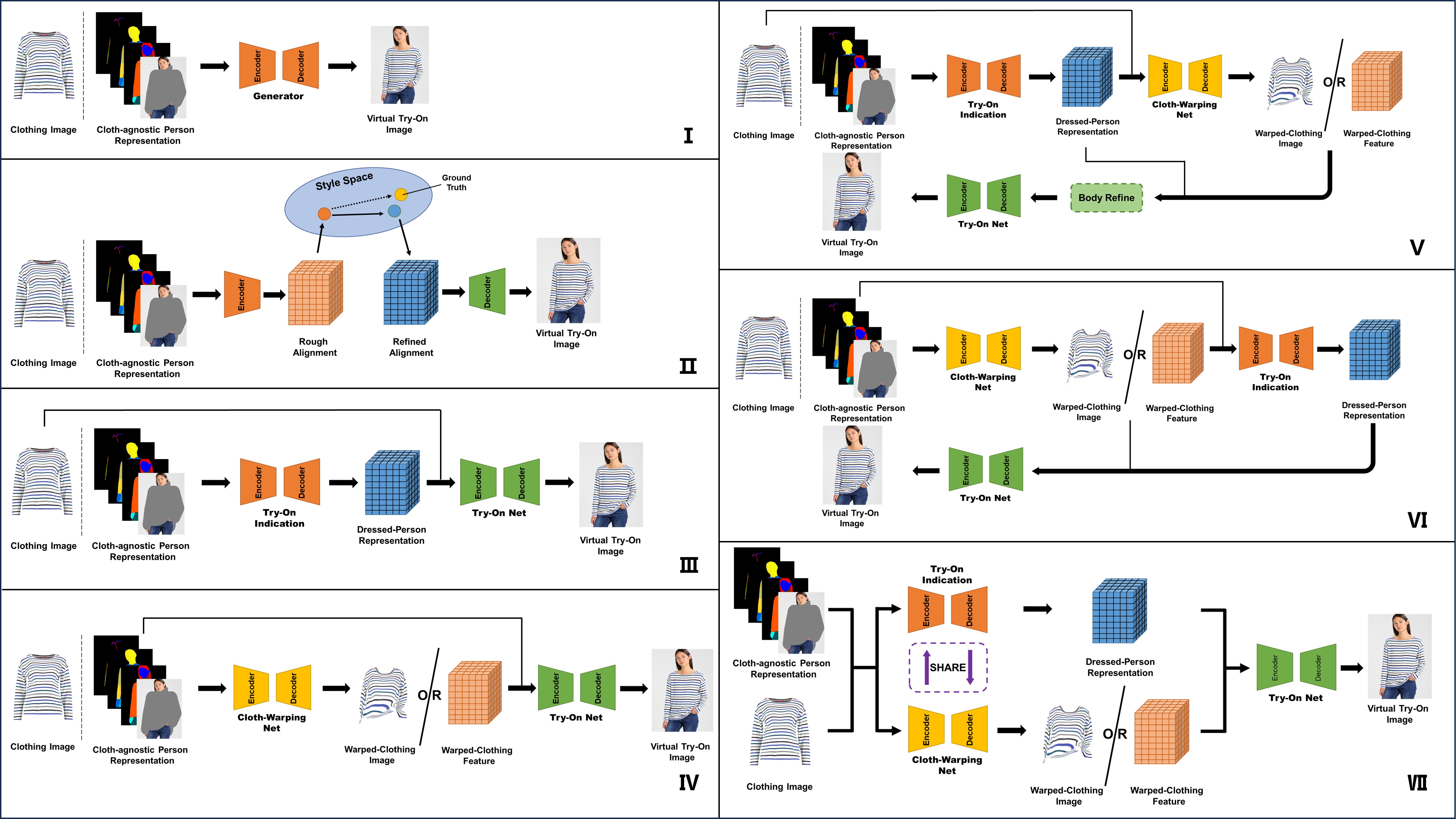}
\caption{Basic pipelines of image-based virtual try-on. Pipelines \Rmnum{1} and \Rmnum{2} are both single-stage approaches, where pipeline \Rmnum{1} utilizes a single generator to directly generate the try-on image, while pipeline \Rmnum{2} aligns features in the feature domain before generating the try-on image. Pipelines \Rmnum{3} and \Rmnum{4} are both two-stage pipelines, where the former utilizes person representation as the bridge while the later uses warped clothing. 
Pipelines \Rmnum{5} and \Rmnum{6} are three-stage pipelines, which differ in the order of Try-On Indication and Cloth Warping. 
Pipeline \Rmnum{7} is an improvement over \Rmnum{5} and \Rmnum{6}, which simultaneously performs Try-On Indication and Cloth Warping.
}
\label{totalmodel}
\end{figure*}

\section{Technical Review and Discussion}\label{Technical review and discussion}
\subsection{Overview}\label{Overview}
Image-based virtual try-on can be regarded as a conditional image generation task that uses in-shop clothing image $I_c$ and person image $I_p$ as raw data, and pre-processes the raw data as conditioned information to guide the model for generating try-on images $I_{try-on}= G(I_p, I_c)$.
Three key modules are usually involved in image-based virtual try-on: 
\begin{itemize}
    \item {\bf{Try-On Indication}} aims to provide a prior for guiding the deformation of clothing in the Cloth Warping module and the fusion of clothing and body in the Try-On module. It usually takes a combination of person body representations (e.g., semantic information \citep{Seg_VITON-HD,Seg_VTNFP_SwapNet}, Densepose \citep{Densepose}, Openpose \citep{pose_VITON,Pose_VITON-HD(Openpose)} and so on) as input, and predicts the spatial structure of person body under the try-on state.
    \item {\bf{Cloth Warping}} transforms the clothing image to the spatial distribution under the try-on state. The inputs of this module are clothing images and person body features such as cloth-agnostic person representation or dressed person representation obtained in the module of Try-On Indication. Via warping methods such as TPS \citep{TPS}, STN \citep{STN}, and FlowNet \citep{FlowNet} that transform the spatial positions of pixels/feature points, the output this module could be warped clothing images or deformed clothing features. 
    \item {\bf{Try-On module}} generates the final try-on image by fusing the person body and clothing features. Interpolation or generative networks are designed for this module, and the output image should meet the following requirements: 1) the clothing within the try-on area should be clear and natural, 2) the content outside the try-on area (excluding the original clothing that is planned to take off) should remain unchanged, and 3) there should be a correct semantic relationship between the new clothing and the person body.
\end{itemize}

It is worth noting that the above three steps are not necessarily present at the same time, and there is no strict order. Table \ref{big_table} summarizes representative methods and we will discuss the key designs in the following subsections. The statistics of image-based virtual try-on methods are summarized from the literature listed in Table \ref{big_table} and the following methods \citep{RCW_CP,PVT_CP,stn_2,stn_3,stn_6,stn_7,tps_1,tps_10,tps_11,tps_12,tps_13,tps_14,tps_15,tps_16,tps_17,tps_18,tps_2,tps_3,tps_4,tps_5,tps_6,tps_7,tps_8,tps_9,tezhengyu_1,tezhengyu_2,PASTA-GAN-plus,tezhengyu_5,flow_1,flow_2,flow_4,flow_5,multimodal-vton,2023_vton_FICE,2023_vton_DOC-vton, 2023_vton_PL-VTON, 2023_vton_Fashiontex, 2023_vton_DM-VTON, 2023_vton_SSAFN, 2023_vton_Dreampaint,2023_vton_VTON-IT}.


\begin{figure}[htbp]
	\centering
	\subfloat[Teacher-Student \uppercase\expandafter{\romannumeral1}]{
		\includegraphics[width=0.48\linewidth]{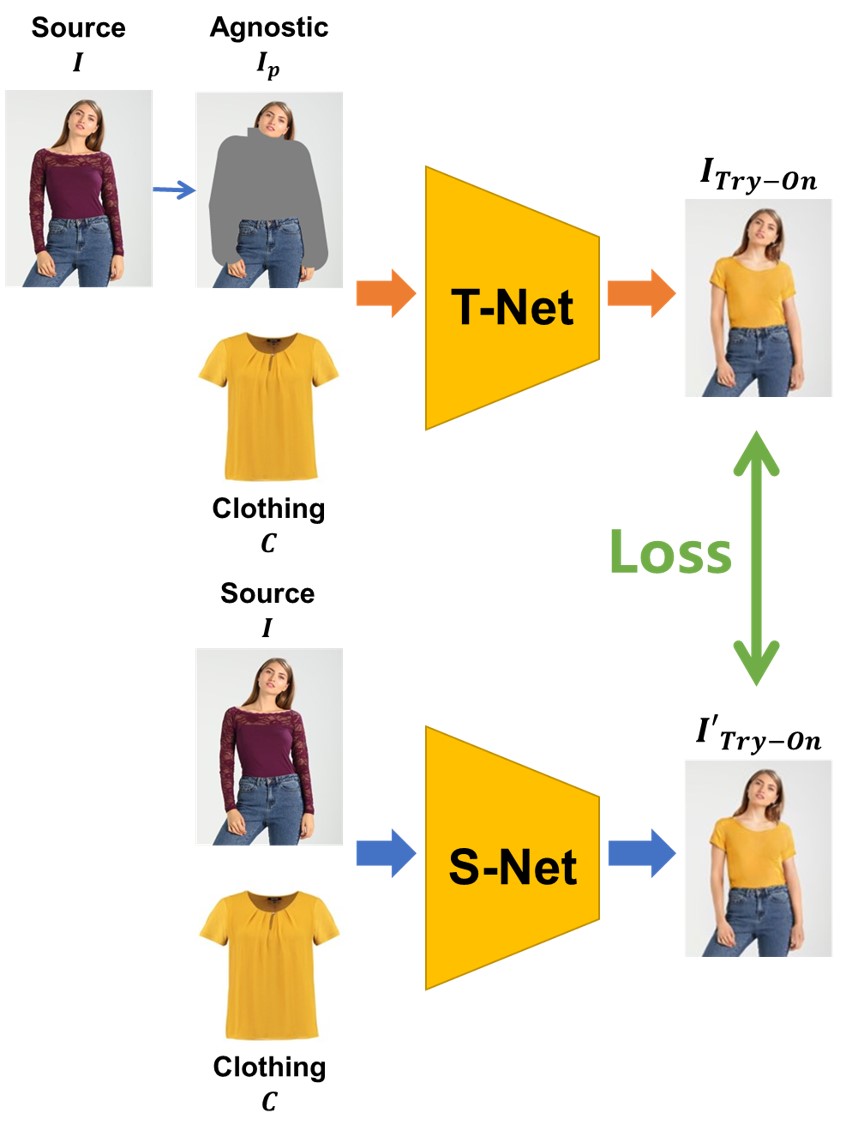}
            \label{Teacher-Stutent_1}} 
        \centering
	\subfloat[Teacher-Student \uppercase\expandafter{\romannumeral2}]{
		\includegraphics[width=0.48 \linewidth]{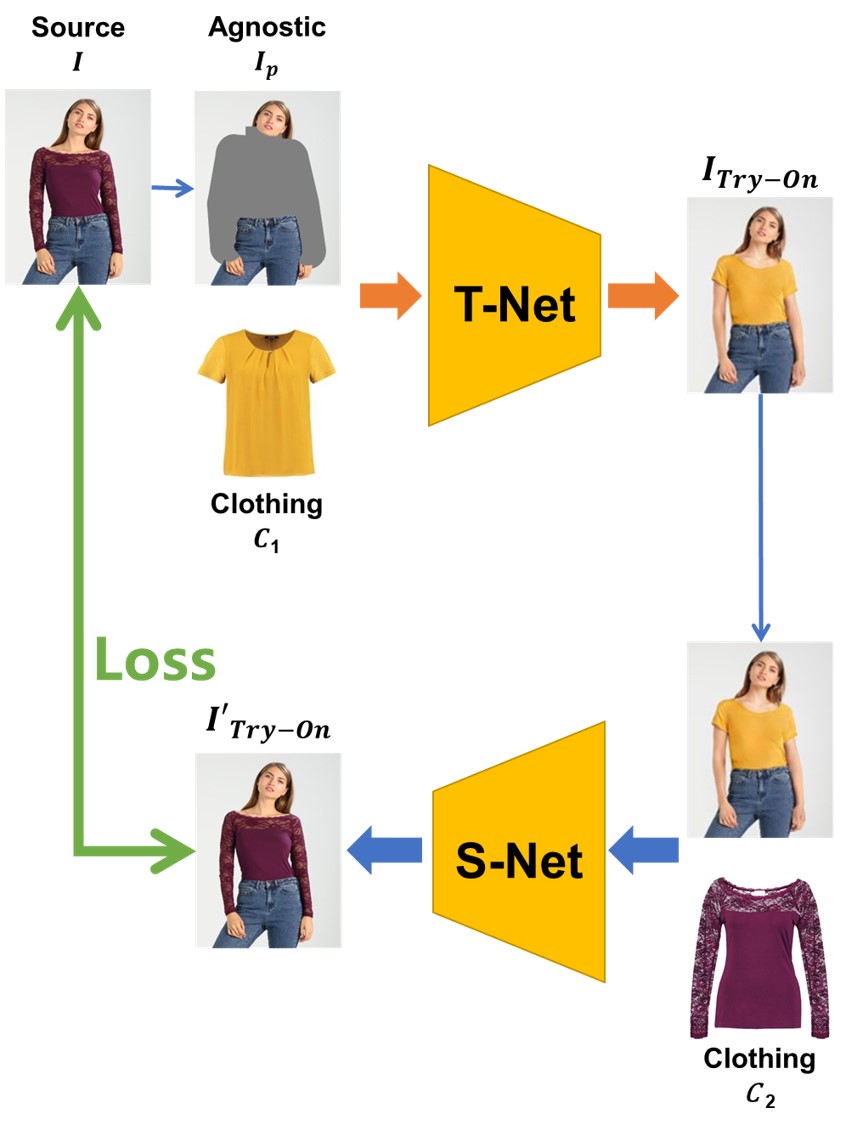}
            \label{Teacher-Stutent_2}}

        \subfloat[Cycle-GAN Structure]{
            \centering
		\includegraphics[width=0.98\linewidth]{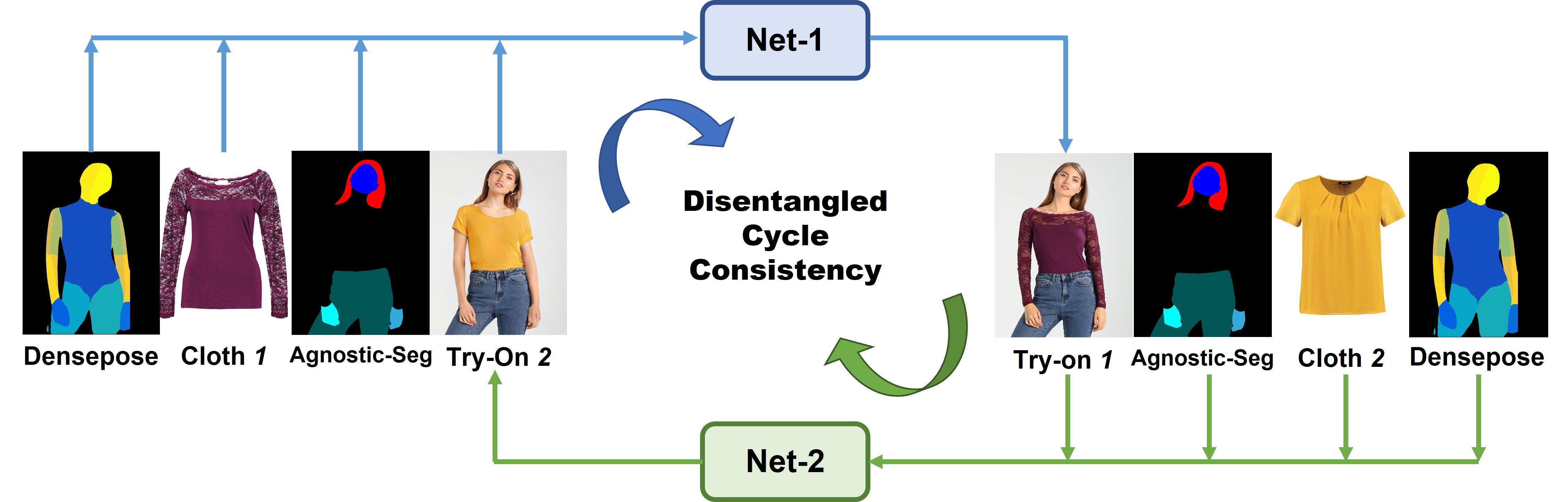}
		\label{DCTON}}
        \caption{Three supplemental structures: (a) Teacher-Student network involving one set of in-shop clothes; (b) Teacher-Student network involving two sets of clothes; (c) Cycle-GAN structure with two sets of clothes.}
    \label{special_network}
 
\end{figure}

\subsection{Pipelines}\label{Pipelines}

In the virtual try-on pipeline, the selection and placement of the aforementioned three modules have a significant impact on the final try-on results. As shown in Fig. \ref{totalmodel}, the basic pipeline structure can be categorized into seven types. Type \Rmnum{1} and \Rmnum{2} are one-stage pipelines and the later one additionally introduces feature alignment. Type \Rmnum{3} and \Rmnum{4} are two-stage pipelines which respectively utilize person representation and warped clothing as intermediate generation for further optimization. The rest types are three-stage pipelines, where type \Rmnum{5} and \Rmnum{6} differ in the order of Try-On Indication and Cloth Warping modules while type \Rmnum{7} simultaneously optimizes these two modules. The pipeline choice of representative methods can be found in Table \ref{big_table}, with no obvious preference in the development trend.

Supplemental to the basic try-on image generation pipeline, Fig. \ref{special_network} shows other structures such as Teacher-Student network  \citep{WUTON,style-flow,PFAFN} and Cycle-GAN  \citep{CycleGAN,SAL-VTON}.  The Teacher-Student architecture is mainly designed for training parser-free try-on network and Fig. \ref{Teacher-Stutent_1} shows the straightforward implementation  \citep{WUTON}. PFAFN and Style-Flow-VTON  \citep{style-flow,PFAFN} further improve it to Fig. \ref{Teacher-Stutent_2} where the synthesized image generated by the teacher network $I_{try-on}$ is used as the input of the student network. Compared with type \Rmnum{1}, type \Rmnum{2} provides more reliable supervision with the ground truth in case that the teacher network generates poor results. Similarly, the adoption of Cycle-GAN  \citep{CycleGAN,SAL-VTON} (Fig. \ref{DCTON}) also shows a strategy for using cycle consistency to enhance the supervision in try-on.
\subsection{Cloth-agnostic Person Representation}\label{Cloth-agnostic Person Representation}
\begin{figure*}[htbp]
\centering
\includegraphics[width=6.2in]{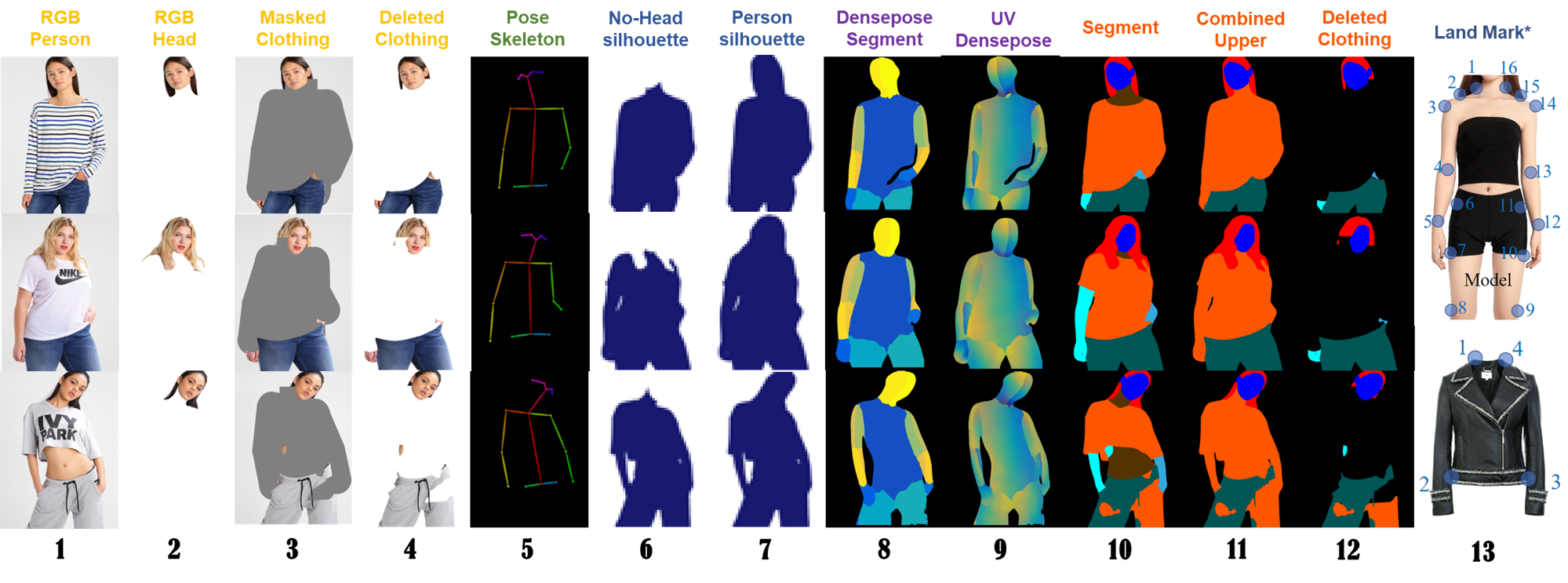}
\caption{Person representation types. Existing representations heavily rely on human parsers to indicate pose, shape or semantic segmentation, which can be categorized into different types: RGB image $\mathcal{P}_{\substack{1,2,3,4}}$, pose $\mathcal{P}_{\substack{5}}$, silhouette $\mathcal{P}_{\substack{6,7}}$, Densepose $\mathcal{P}_{\substack{8,9}}$, semantic segmentation $\mathcal{P}_{\substack{10,11,12}}$, and landmark $\mathcal{P}_{\substack{13}}$. * Subfigures of $\mathcal{P}_{\substack{13}}$ are cited from  \citep{DGP_yuduiqi_ESF}.}
\label{fig:person_representation}
\end{figure*}

In practical scenarios, it is difficult to acquire the triplet (person image $I_s$, in-shop clothing image $C_t$, try-on image $I_t$) for training. Instead, pairs of person image and in-shop clothing image are commonly seen. Therefore, cloth-agnostic person representation is studied to remove the clothing that is planned to take off, and constitute the triplet (cloth-agnostic person representation $I_{agnostic}$, in-shop clothing image $C_t$, person image $I_t$) for supervised training. Existing works employ human parser to extract representations such as pose, shape and semantic distribution from the person body. As shown in Fig. \ref{fig:person_representation}, the representations can be categorized into different types: RGB image $\mathcal{P}_{\substack{1,2,3,4}}$, pose keypoints $\mathcal{P}_{\substack{5}}$, silhouette $\mathcal{P}_{\substack{6,7}}$, Densepose $\mathcal{P}_{\substack{8,9}}$, semantic segmentation $\mathcal{P}_{\substack{10,11,12}}$, and landmark $\mathcal{P}_{\substack{13}}$.

RGB image $\mathcal{P}_{\substack{1,2,3,4}}$ provides pixel-level features that can better preserve the identity information of individuals and the content outside the try-on area. $\mathcal{P}_{\substack{1}}$ shows the original person image which is disturbed by the original clothes. $\mathcal{P}_{\substack{2,3,4}}$ can be acquired via human parsing  \citep{Pose_VITON-HD(Openpose),Seg_VITON-HD,parsing_survey}, where $\mathcal{P}_{\substack{2}}$ contains the head information, $\mathcal{P}_{\substack{3}}$ masks the try-on related area using large areas of gray pixels, and $\mathcal{P}_{\substack{4}}$ deletes the masked area with background color. Comparatively, $\mathcal{P}_{\substack{3}}$ contains some clues about pose and shape while $\mathcal{P}_{\substack{4}}$ completely removes the try-on related body area. 

Pose features $\mathcal{P}_{\substack{5}}$ \citep{pose_VITON,Pose_VITON-HD(Openpose)} estimate the positions of 18 key points of the body, which provides an important condition for generating person body images and directly affects clothing deformation.

Silhouette $\mathcal{P}_{\substack{6,7}}$ refers to the body contour that contains rough pose and shape information. Cloth-agnostic representation was initially proposed by VITON  \citep{VITON} via downsampling to a lower resolution to make the contour unclear about the original clothes.

To further separate the shape representation from original clothes, SP-VITON  \citep{sp-VITON} initially utilizes Densepose  \citep{Densepose} to estimate the body shape under clothes.
Densepose also provides other formats such as semantic parsing  $\mathcal{P}_{\substack{8}}$ and UV map coordinates $\mathcal{P}_{\substack{9}}$ corresponding to the 3D model.

Semantic segmentation $\mathcal{P}_{\substack{10,11,12}}$ is used to provide knowledge about the try-on area in virtual try-on. However, the contour of semantic distribution $\mathcal{P}_{\substack{10}}$ indicates the original clothes and has negative impacts on trying on new clothes. Consequently, a combination of clothes and related skin $\mathcal{P}_{\substack{11}}$ aims to eliminate the prior influence of the original clothing style. Furthermore, $\mathcal{P}_{\substack{12}}$ completely removes the contour of the original clothing, retaining only the semantic regions unrelated to the try-on area. Although totally getting rid of the original clothes, the complete deletion also looses body shape priors.

Landmark $\mathcal{P}_{\substack{13}}$ provides explicit semantic information for shape alignment. It guides clothing deformation by constraining the distance between clothing landmarks and corresponding body landmarks. SAL-VTON \citep{SAL-VTON} makes improvements based on HR-Net \citep{HRNet} and proposes a landmark estimation network.

The choices of representative methods are shown in Table \ref{big_table}, where pose keypoints are most commonly used and silhouette falls behind the other types. Additionally, there is no obvious preference towards representation types and a combination of several types could also be used.
\subsection{Try-On Indication}\label{Try-On Indication}
Given the cloth-agnostic person representation and target clothing information as input, the Try-On Indication network (i.e., the orange module in Fig. \ref{totalmodel}) is used to predict the dressed person representation, which directly influences the generation of the final try-on image. The architectures of this module and corresponding constraint losses for representative methods are shown in Table \ref{big_table}, where the encoder-decoder structure is the mainstream framework for Try-On Indication. It encodes cloth-agnostic person representation and decodes the representation of dressed person under the condition of the target clothing.

VITON-series works  \citep{VITON, CP-VTON, sp-VITON,vtnfp,RCW_CP,PVT_CP,LA-VTON,CP-VTON+} such as VITON \citep{VITON} and CP-VTON \citep{CP-VTON} use the U-Net \citep{U-Net} architecture and person representation $\mathcal{P}_{\substack{2,5,6}}$ to directly generate coarse try-on images, but sometimes generate blurry trunk and misalignment of the clothing edges. To overcome this issue, CIT \citep{clothflow} added the Transformer \citep{transformer} structure to the basis of CP-VTON, but still cannot totally solve this problem.

Instead of predicting pixels of rough try-on images, ClothFlow  \citep{clothflow_flownet} uses U-Net to estimate the semantic distribution of dressed persons. Some other works \citep{ACGPN,LM-VITON,zflow,VITON-HD,OVNet} such as ACGPN \citep{ACGPN} and LM-VTON  \citep{LM-VITON} follow the same strategy and combine the person representation with $\mathcal{P}_{\substack{2,5,6,7,8,11}}$, further improving the try-on effects. 

Instead of single-step prediction, multiple steps could also improve the generation quality.
UVTON first takes $\mathcal{P}_{\substack{4}}$ and clothing images as input, generates RGB images of ten body areas through ten generators, and then uses this set of images and $\mathcal{P}_{\substack{4,9}}$ to generate coarse try-on images. ACGPN  \citep{ACGPN} proposes a Try-On Indication composed of two serial GAN generators. The first generator uses $\mathcal{P}_{\substack{5,11}}$ features and clothing images to predict the semantic distribution. Then the predicted semantic distribution, pose keypoints $\mathcal{P}_{\substack{5}}$ and clothing images are input into the second generator to predict the mask of the clothing area. O-VITON  \citep{o-VTON} introduces a shape generation network, which first encodes the shapes of different body parts, and then inputs the encoding values and UV Densepose $\mathcal{P}_{\substack{9}}$ into the generator to generate the semantic map.

Indeed, the Try-On Indication module and the Clothing Warping module are closely related and affect each other. 
HR-VTON  \citep{HR-VTON} inputs clothing images, clothing Masks and person representation $\mathcal{P}_{\substack{8,13}}$ at the same time, and simultaneously generates warped clothes and person's semantic distribution. The warping path and the semantic prediction path can keep communication through the Fusion Block.

As shown in Table \ref{big_table}, the commonly-used loss for constraining Try-On Indication module mainly involve mask L1/L2, cross entropy loss, adversarial loss and perceptual loss. Specifically, mask L1/L2 is used to constrain the clothing mask, cross entropy loss facilitates the segmentation prediction and perceptual loss is designed for the generated RGB image. Adversarial loss is adopted for GAN-based generation methods. Both of Focal loss  \citep{vtnfp_focalloss} and LSGAN loss  \citep{LSGANloss} are designed to constrain the generation of segmentation.
\subsection{Cloth Warping}\label{Cloth Warping}
Shown as the yellow module in the pipeline (Fig. \ref{totalmodel}, Cloth Warping module aims to transform the spatial distribution of the clothing image/feature to match the body. The mainstream deformation methods can be classified into: Thin Plate Spline (TPS), Spatial Transformation Network (STN), Flow Net and Implicit transformation. The choices of representative methods and corresponding losses are illustrated in Table \ref{big_table}.
\begin{figure*}[htbp]
	\centering
	\subfloat[TPS]{
		\includegraphics[width=0.4\linewidth]{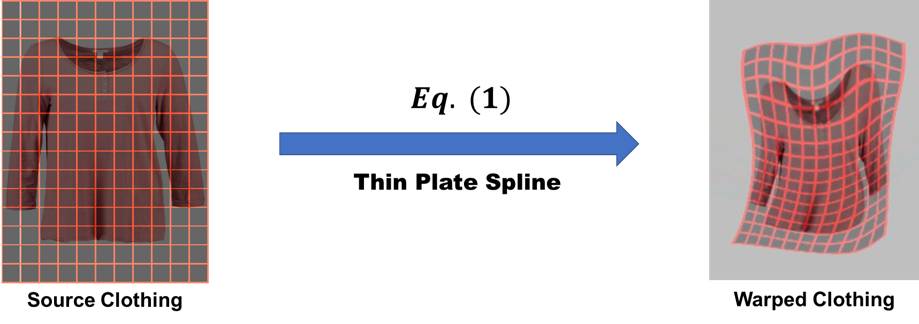}
            \label{TPSshiyitu}} 
        \hspace{1cm}
	\subfloat[STN]{
		\includegraphics[width=0.4\linewidth]{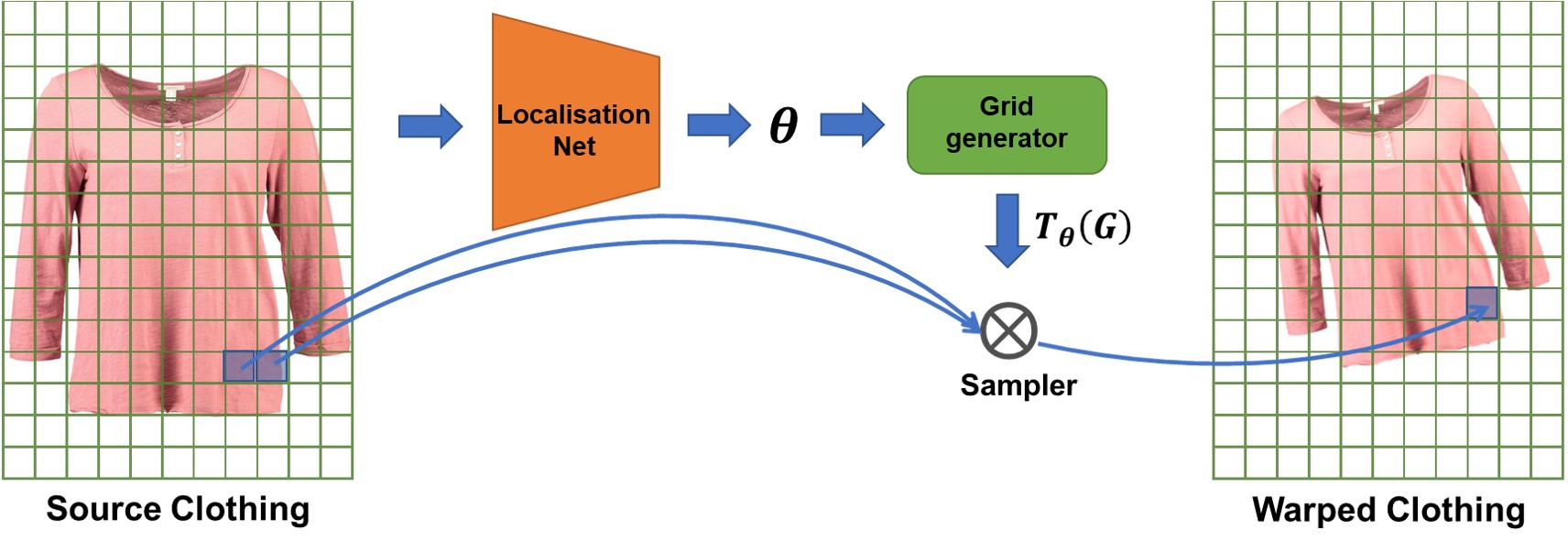}
            \label{STNshiyitu}} 
        \\
	\subfloat[Flow]{
		\includegraphics[width=0.4\linewidth]{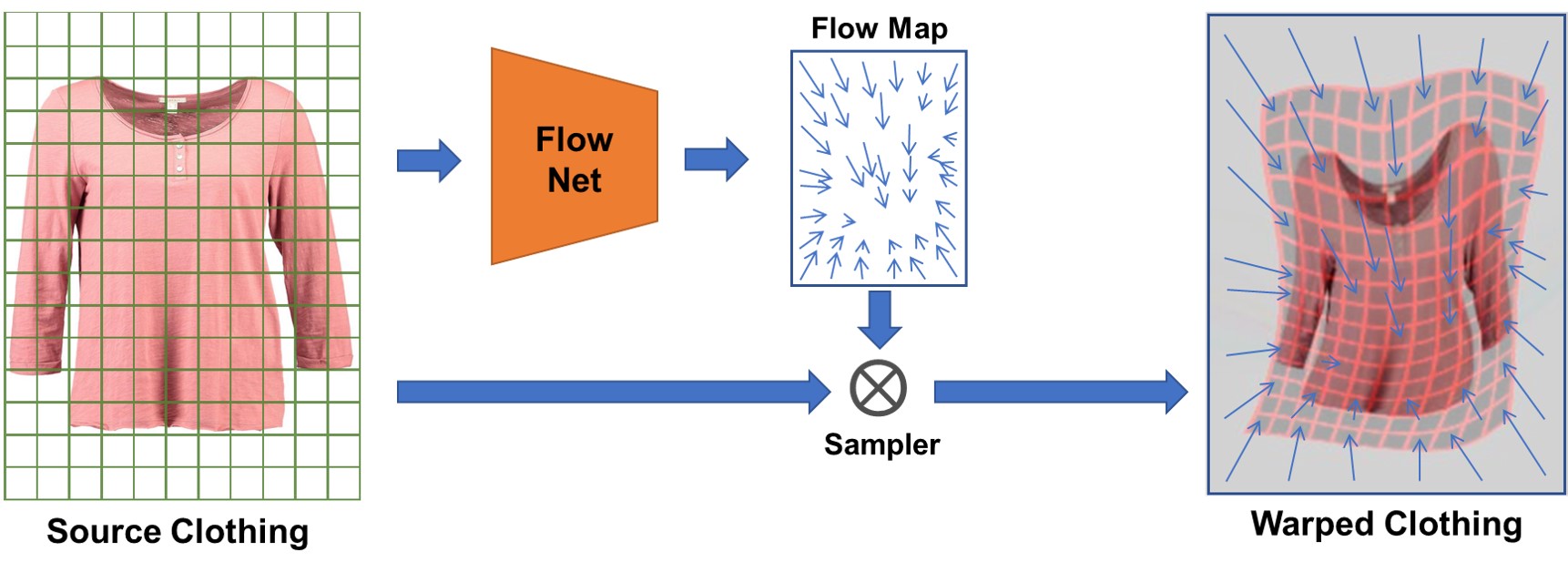}
            \label{Flow_netshiyitu}} 
        \hspace{1cm}
	\subfloat[Number Of Works / Year]{
		\includegraphics[width=0.4\linewidth]{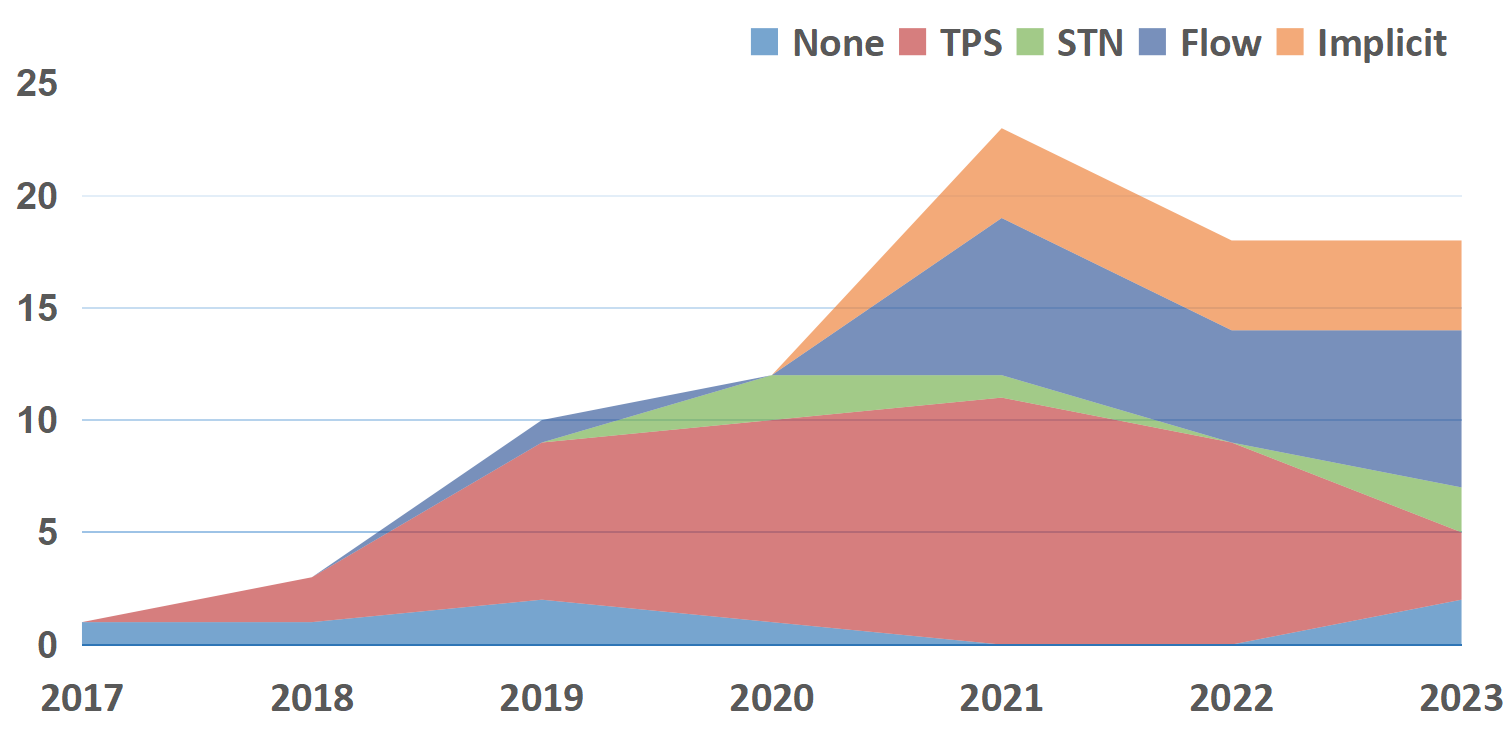}
            \label{qvshitu}}

        \caption{Illustration for several spatial transformation approaches: (a) Thin Plate Spline (TPS), (b) Spatial Transformation Network (STN) and (c) Flow. (d) shows the usage statistics of existing methods.
        }
    \label{clothing-warp}
 
\end{figure*}

\subsubsection{Thin Plate Spline}\label{Thin Plate Spline}
Thin Plate Spline (TPS) interpolation method use the deformation of a thin steel plate to simulate 2D deformation, which is utilized by most methods for warping clothes. 
Suppose there are N control points (e.g., the nodes of image grid shown in Fig. \ref{TPSshiyitu}), we denote the original coordinates as $T={(t_1,t_2,\ldots\ldots,t_n)}^T$ and target coordinates as $Y={(y_1,y_2,\ldots\ldots,y_n)}^T$.
Take one point $y_i$ as an example, its interpolation under TPS transformation can be formulated as:
\begin{equation}
\begin{aligned}
\Phi(y_i)=c+\mathbf{A}\mathbf{y_i}+\sum_j^N (w_j U(\lVert \mathbf{y_i}-\mathbf{t_j} \rVert))
\end{aligned}
\label{Eq:TPS}
\end{equation}
\noindent where $c$ and $A$ simulate linear transformation, $U(\cdot)$ represents a collection of radial basis functions, and $w_i$ is the corresponding weighting parameter. 
Please refer to  \citep{TPS} for more details.

Given a series of control points and their target positions, target image under TPS interpolation could be computed. However, for image-based virtual try-on, the lack of ground truth for target positions becomes an obstacle for TPS-based methods. VITON \citep{VITON} utilizes shape context matching  \citep{shape_tps} to estimate target positions based on the context relationship of the clothing mask and the predicted warped mask. CP-VTON  \citep{CP-VTON} designs a Geometry Matching Module (GMM, one type of Spatial Transformer Network) to feed $\mathcal{P}_{\substack{2,5,6}}$ and learns TPS parameters for clothing warping. WUTON \citep{WUTON} uses TPS transformation at the feature level, which can enhance the diversity of clothing warping but has a shortage of image clarity. 

With the help of spatial transformation network (STN, which will be introduced next), further trials have been made to better constrain TPS transformation: 1) Regularization. To constrain the warping of clothing, LA-VTON \citep{LA-VTON} proposes the constraint of grid interval consistency, which limits the distance between adjacent nodes; CP-VTON+ \citep{CP-VTON+} introduces a regularization term that calculates the sum of distance changes for N nodes of grid; ACGPN \citep{ACGPN} proposes a second-order difference constraint to prevent exaggerated deformation; DCTON \citep{DCTON} uses a homography matrix $H$ to reduce the change of the transformation matrix. 2) Auxiliary information. LM-VTON \citep{LM-VITON} designs a landmark-based method for fine-grained constraint. VITON-HD \citep{VITON-HD} additionally uses $\mathcal{P}_{\substack{3,5}}$ and the clothing region Mask, while C-VTON \citep{C-VTON} uses Densepose features alone as the input of GMM. 3) New model. CIT \citep{clothflow} replaces the association layer in GMM with Transformer structure to better calculate the spatial correlation between the person pose and the clothing.
\subsubsection{Spatial Transformation Network}\label{Spatial Transformation Network}
Spatial Transformation Network (STN)  \citep{STN} can be embedded into a layer of the network, and consists of three main parts as shown in Fig. \ref{STNshiyitu}: 
1) Localisation Network is a simple regression network which contains several convolutional layers to regress $N$ variables (for example, a $2\times3$ matrix with 6 variables in the case of affine transformation).
2) Grid Generator is responsible for calculating the coordinates of the target image corresponding to the coordinates of the original image through matrix operations with $T_\theta(G)$.
3) Sampler samples the original image according to the coordinates in $T_\theta(G)$ and copies the pixels from original image to target image.

To simulate complex garment deformation, single STN usually cooperates with TPS, such as GMM in CP-VTON  \citep{CP-VTON}. Multiple STNs are another way to capture complex deformations by iteratively performing multiple transformations (e.g., affine transformation). OVNet \citep{OVNet} decomposes the warping of clothing into multiple small warps with excellent results, which proposes a multi-warp clothing generator to obtain $k$ warping results. The $k$ deformed images are connected in the channel scale and input into the Try-On module together to facilitate the generation of try-on image.
\subsubsection{Flow Estimation}\label{Flow Estimation}
Flow indicates the offset of pixel or feature before and after transformation. Let $(u_x, u_y)$ denote the offset, the value at target position $(x,y)$ can be sampled at $(x-u_x, y-u_y)$ in the original distribution, and the non-integer coordinates are interpolated by bilinear interpolation. Flow estimation methods for cloth warping can be classified in terms of prediction target such as pixel and feature or prediction steps such as single layer or multiple layers.

Dior \citep{DiOr} employs a single-layer flownet to deform the clothing features, where GFFE (Global Flow Field Estimator) proposed by \cite{GFLA} is used to predict flow map and soft mask. To capture more 3D changes of human posture deformation, 3D-GCL \citep{3D-GCL} uses  flownet  \citep{FlowNet} to create SMPL \citep{SMPL} flows as ground truth to constrain the flow map, which results in better visual outcomes on the Deep Fashion dataset. \reviseMark{RP-VTON \citep{RP-VTON} utilizes landmarks to control cloth warping, which is novel in achieving controllable try-on.}

A single-layer flownet applies sampling just at one level, while a multi-layer flownet conducts sampling from coarse to fine across multiple feature levels. ClothFlow \citep{clothflow_flownet} adopts a dual-path multi-layer decoder, which performs flow map calculation and warping at the feature level in each layer. Finally, the pixel-level flow map is output in the last layer. 
Similar to the structure of flownet in Clothflow, PFAFN \citep{PFAFN} uses more human parsing features $\mathcal{P}_{\substack{5,8,13}}$ to guide the estimation of Flownet. To extract more flow features, ZFlow \citep{zflow} procures flow maps of identical sizes from various depths within U-net via interpolation, which are then consolidated into a single flow map. Intuitively, this is a pixel-by-pixel selection process, which determines the total flow rate by selecting (allowing or excluding) the pixel flow estimation of different radial neighborhoods (for multiple scales). 
In order to prevent excessive deformation caused by the high degree of freedom of flownet, total variation loss is added to the loss function as a regularization term. 

There are also some other interests in flow estimation. HR-VTON \citep{HR-VTON} and SDAFN \citep{SDFN} put emphasis on the coupling between body features and clothing.
To counteract the potential image distortion resulting from flownet's excessively focus on local alignment, Flow-Style-VTON \citep{style-flow} uses StyleGAN network to capture global deformation information to generate a coarse flow map, and then refines it locally to output target flow map. To achieve the refinement of flow estimation, multi-flow is gaining increasing popularity. SDAFN \citep{SDFN}, SAL-VTON \citep{SAL-VTON} and GP-VTON \citep{GP-VTON} construct a warping network from coarse to fine. Compared with the general flownet, they use multiple flow maps to estimate multiple factors such as posture, body shape and mask of clothing. 
\subsubsection{Implicit Transformation}\label{Implicit Transformation}
In O-VITON \citep{o-VTON}, TryOnGAN \citep{Tryongan_yuduiqi}, and DGP \citep{DGP_yuduiqi_ESF}, no explicit spatial transformation method is adopted, but after deep encoding, the clothes is aligned to the target posture in the feature space. O-VITON, as an early work adopting implicit transformation, simply diffuses clothing features in the corresponding person body region. TryOnGAN and DGP are recent works, which adopt the alignment in the feature space of StyleGAN \citep{stylegan2,stylegan1} to deform clothing to the target posture, achieving seamless coverage of the person body.

As shown in Fig. \ref{qvshitu}, there has been a diversification in clothing warping methods in recent years. Single-STN methods are unable to handle natural deformation. Apart from a few works employing Multi-STN, most methods use a combination of STN and TPS, and they are classified under the TPS category. Influenced by the GMM framework, TPS transformation is the most popular method. TPS transformations have limitations in terms of degrees of freedom for deformation. In order to achieve more diverse clothing warping results, flow estimation has become a breakthrough in deformation performance. At the cost of greater computational overhead, flownet has achieved stronger deformation capabilities, and its usage is on the rise. Implicit transformation methods deform garments in a generative manner, further enhancing the diversity of garment deformation but also posing challenges in controlling clothing content. \reviseMark{With the development of image generation technology under the diffusion framework, implicit transformation methods become promising.}

Two kinds of loss for clothing warping are commonly adopted: 1) L1 loss and perceptual loss to supervise the warping with ground truth; 2) Regularization to alleviate exaggerated deformation, such as grid interval consistency \citep{LA-VTON}, second-order-difference constraint \citep{ACGPN,VITON-HD,GP-VTON}, landmark loss \citep{LM-VITON,SAL-VTON}, grid regularization \citep{CP-VTON+,clothflow}, homography matrix regularization \citep{DCTON} and TV (total variation) norm \citep{clothflow_flownet,zflow,HR-VTON}.
\subsection{Try-On}\label{Try-On}
Try-On module is the final stage of the try-on pipeline, which combines the clothing and person information obtained in the previous modules to generate the final try-on image. Therefore, it directly affects the quality of the output image. At present, there are two kinds of methods in the literature, one is to use mask for combining the person image and the warped clothing image \citep{VITON,CP-VTON,sp-VITON,CP-VTON+,clothflow,RCW_CP,PVT_CP}, and the other is using the generation network with the clothing and person features \citep{ACGPN,o-VTON,LM-VITON,WUTON,OVNet,DCTON,PFAFN,VITON-HD,DiOr,zflow,Tryongan_yuduiqi,style-flow,DGP_yuduiqi_ESF,HR-VTON,swapnet,u-vton,LA-VTON,clothflow_flownet,SDFN}. Comparatively, the latter one obtains better generation performance at the cost of computation, which is the current mainstream approach.

{\bf{Mask Combination: }}Such methods 
adopt the mask combination to generate the final try-on image, which is formulated as $I=I_{coarse} \cdot (1-M)+C \cdot M$. The input is the coarse try-on image $I_{coarse}$, the warped clothing image $C$ and the mask $M$ representing the clothing region of the dressed person. Essentially, the warped clothing is covered in the area represented by the $M$ in the $I_{coarse}$. This is a relatively simple and direct method, the advantage of which is that it will not increase the number of network parameters, but the disadvantage is that the quality of the final try-on image $I$ depends entirely on the correctness of the clothing warping and the coarse try-on image where misaligned regions cause artifacts.

{\bf{Generation: }}Generation-based methods 
all use a generator to design the Try-On module. Among them,  \citep{WUTON,OVNet,zflow,PFAFN,style-flow,swapnet,vtnfp} use U-Net as the generator. Besides the try-on image as the generation target, VTNFP \citep{vtnfp} uses U-Net to generate the try-on image and the combination mask at the same time, where combination mask is used to optimize the clothing details of the try-on image. The U-Net in ZFlow \citep{zflow} additionally generates representation features $\mathcal{P}_{\substack{8,9,10}}$ and calculates the loss between these three features and the ground truth to assist the training of the Try-On network.

For high-resolution virtual try-on \citep{VITON-HD, HR-VTON}, misalignment regions between the clothing and the person body become more pronounced, and special designs for the generator are expected to eliminate the misalignment between clothing and body. VITON-HD \citep{VITON-HD} uses ALIAS ResBlock in the decoder part of the generator, and retains better spatial distribution information of clothing and person features through SPADE \citep{SPADE} normalization when generating try-on images. Moreover, it explicitly calculates the misalignment area between person and clothing in the residual block, guiding the network to fill the texture features of clothing into the misalignment area. In HR-VITON \citep{HR-VTON}, the dressed person semantic distribution and the warped clothing image are aligned first, eliminating the misaligned areas. To constrain the complex process of semantic map processing, a ``discriminator rejection method" was proposed to eliminate the low-quality semantic information map during adversarial training.

\begin{table*}[t!]
\caption{Key properties of datasets. Popularity refers to the number of usage from existing methods.
}
\label{table_dataset}
\centering
\resizebox{1\textwidth}{!}{
\setlength\tabcolsep{2.5pt}
\renewcommand\arraystretch{1}
\begin{tabular}{cccccccccccccc}
\toprule
\textbf{\begin{tabular}[c]{@{}c@{}}Dataset\end{tabular}} & 
\textbf{\begin{tabular}[c]{@{}c@{}}Year\end{tabular}} &
\textbf{\begin{tabular}[c]{@{}c@{}}Clothing\\ Image\end{tabular}}& \textbf{\begin{tabular}[c]{@{}c@{}} Women\end{tabular}} & \textbf{\begin{tabular}[c]{@{}c@{}} Men\end{tabular}} & \textbf{\begin{tabular}[c]{@{}c@{}}Upper\\body\end{tabular}} & \textbf{\begin{tabular}[c]{@{}c@{}}Lower\\body\end{tabular}} & \textbf{\begin{tabular}[c]{@{}c@{}}Full\\body\end{tabular}} & \textbf{\begin{tabular}[c]{@{}c@{}}Multi\\Pose\end{tabular}} & \textbf{\begin{tabular}[c]{@{}c@{}}Human\\Parsing\end{tabular}} & \textbf{\begin{tabular}[c]{@{}c@{}}\reviseMark{Shop}\\ \reviseMark{Wild}\end{tabular}} & 
\textbf{\begin{tabular}[c]{@{}c@{}}Resolution\end{tabular}} & \textbf{\begin{tabular}[c]{@{}c@{}}Quantity\\Train/Test \end{tabular}} &
\textbf{\begin{tabular}[c]{@{}c@{}}Popu-\\larity\end{tabular}} 
\\\midrule

VITON & 2018 & \Checkmark & \Checkmark & \ding{55} &  \Checkmark &  \ding{55} & \ding{55} & \ding{55} & \Checkmark  & \reviseMark{S}& \begin{tabular}[c]{@{}c@{}}$256\times192$ \\ $1024\times768$ \end{tabular} & \begin{tabular}[c]{@{}c@{}}14,221/2,032 \end{tabular} & 48 \\
VITON-HD & 2021 & \Checkmark & \Checkmark & \ding{55} &  \Checkmark &  \ding{55} & \ding{55} & \ding{55} & \ding{55}  & \reviseMark{S}& \begin{tabular}[c]{@{}c@{}}$1024\times768$ \end{tabular} & \begin{tabular}[c]{@{}c@{}}11,647/2,032 \end{tabular} & 17\\
DeepFashion & 2016 & \ding{55} & \Checkmark & \Checkmark &  \Checkmark &  \Checkmark & \Checkmark & \Checkmark & \Checkmark  & \reviseMark{S}& \begin{tabular}[c]{@{}c@{}}$1101\times750$ \end{tabular} & \begin{tabular}[c]{@{}c@{}}52,712/- \end{tabular} & 13 \\
MPV & 2019 & \Checkmark & \Checkmark & \ding{55} &  \Checkmark &  \ding{55} & \ding{55} & \Checkmark & \Checkmark  & \reviseMark{S}& \begin{tabular}[c]{@{}c@{}}$256\times192$ \end{tabular} & \begin{tabular}[c]{@{}c@{}}52,236/10,544 \end{tabular} & 12\\
Dress Code & 2022 & \Checkmark & \Checkmark & \Checkmark &  \Checkmark &  \Checkmark & \Checkmark & \ding{55} & \Checkmark  & \reviseMark{S}& \begin{tabular}[c]{@{}c@{}} $1024\times768$ \end{tabular} & \begin{tabular}[c]{@{}c@{}}48,392/5,400 \end{tabular} & 11 \\
\reviseMark{SHHQ} & \reviseMark{2022} & \reviseMark{\ding{55}} & \reviseMark{\Checkmark} & \reviseMark{\Checkmark} &  \reviseMark{\ding{55}} &  \reviseMark{\ding{55}} & \reviseMark{\Checkmark} & \reviseMark{\ding{55}} & \reviseMark{\ding{55}}  & \reviseMark{S}& \begin{tabular}[c]{@{}c@{}}\reviseMark{$1024\times512$} \end{tabular} & \begin{tabular}[c]{@{}c@{}}\reviseMark{231,176/-} \end{tabular} & \reviseMark{2} \\
\reviseMark{UPT} &\reviseMark{2021} & \reviseMark{\ding{55}} & \reviseMark{\Checkmark} & \reviseMark{\Checkmark} &  \reviseMark{\Checkmark} &  \reviseMark{\Checkmark} & \reviseMark{\Checkmark} & \reviseMark{\ding{55}} & \reviseMark{\Checkmark}  & \reviseMark{S}& \begin{tabular}[c]{@{}c@{}}\reviseMark{$512\times320$} \end{tabular} & \begin{tabular}[c]{@{}c@{}}\reviseMark{27,139/6,115} \end{tabular} & \reviseMark{2} \\
ESF & 2022 & \ding{55} & \Checkmark & \Checkmark &  \Checkmark &  \ding{55} & \Checkmark & \ding{55} & \ding{55}  & \reviseMark{S}& \begin{tabular}[c]{@{}c@{}}$512\times512$ \end{tabular} & \begin{tabular}[c]{@{}c@{}}170,000/10,000 \end{tabular} & 1 \\
\reviseMark{StreetTryOn} &\reviseMark{2024} & \reviseMark{\ding{55}} & \reviseMark{\Checkmark} & \reviseMark{\Checkmark} &  \reviseMark{\Checkmark} &  \reviseMark{\ding{55}} & \reviseMark{\Checkmark} & \reviseMark{\ding{55}} & \reviseMark{\Checkmark} & \reviseMark{W} & \begin{tabular}[c]{@{}c@{}}\reviseMark{$512\times320$} \end{tabular} & \begin{tabular}[c]{@{}c@{}}\reviseMark{12,364/2,089} \end{tabular} & \reviseMark{1} \\
\reviseMark{LH-400K} &\reviseMark{2024} & \reviseMark{\ding{55}} & \reviseMark{\Checkmark} & \reviseMark{\Checkmark} &  \reviseMark{\Checkmark} &  \reviseMark{\ding{55}} & \reviseMark{\Checkmark} & \reviseMark{\ding{55}} & \reviseMark{\Checkmark}  & \reviseMark{W}& \begin{tabular}[c]{@{}c@{}}\reviseMark{$512\times512$} \end{tabular} & \begin{tabular}[c]{@{}c@{}}\reviseMark{409,270/-} \end{tabular} & \reviseMark{1} \\
\reviseMark{WPose} &\reviseMark{2024} & \reviseMark{\ding{55}} & \reviseMark{\Checkmark} & \reviseMark{\Checkmark} &  \reviseMark{\Checkmark} &  \reviseMark{\ding{55}} & \reviseMark{\Checkmark} & \reviseMark{\ding{55}} & \reviseMark{\Checkmark} & \reviseMark{W} & \begin{tabular}[c]{@{}c@{}}\reviseMark{-} \end{tabular} & \begin{tabular}[c]{@{}c@{}}\reviseMark{-/2,304} \end{tabular} & \reviseMark{1}\\
\reviseMark{WVTON} &\reviseMark{2024} & \reviseMark{\Checkmark} & \reviseMark{\Checkmark} & \reviseMark{\Checkmark} &  \reviseMark{\Checkmark} &  \reviseMark{\ding{55}} & \reviseMark{\Checkmark} & \reviseMark{\ding{55}} & \reviseMark{\Checkmark} & \reviseMark{W} & \begin{tabular}[c]{@{}c@{}}\reviseMark{-} \end{tabular} & \begin{tabular}[c]{@{}c@{}}\reviseMark{-/440} \end{tabular} & \reviseMark{1}\\
\bottomrule
\end{tabular}}
\end{table*}

Previously, only using the Try-On module (without explicit try-on indication and cloth warping) could only generate rough try-on images. However, this issue has been alleviated in the era of diffusion models. TryOnDiffusion \citep{tryondiffusion} has designed a substantial Parallel-UNet, which, after being trained with massive amounts of data, can directly use the original image and pose features to accomplish high-quality try-on. \reviseMark{More recent diffusion-based methods put additional emphasis on attention mechanism \citep{Stable-VITON}, acceleration strategy \citep{CAT-DM}, and image editing techniques \citep{PICTURE,UniHuman}}.

As illustrated in Table \ref{big_table}, besides the commonly-used L1 loss, perceptual loss and adversarial loss, some works further design local constraint (e.g., content preserving loss  \citep{DCTON} and editing-localization loss  \citep{Tryongan_yuduiqi}) and additional semantic constraint (e.g., feature matching loss  \citep{VITON-HD} and attribute loss  \citep{DGP_yuduiqi_ESF}). Diffusion-based  methods usually adopt denoising loss, and \cite{Stable-VITON} additionally introduces attention total variation loss.

\begin{figure*}[t!]
\centering
\includegraphics[width=6.2in]{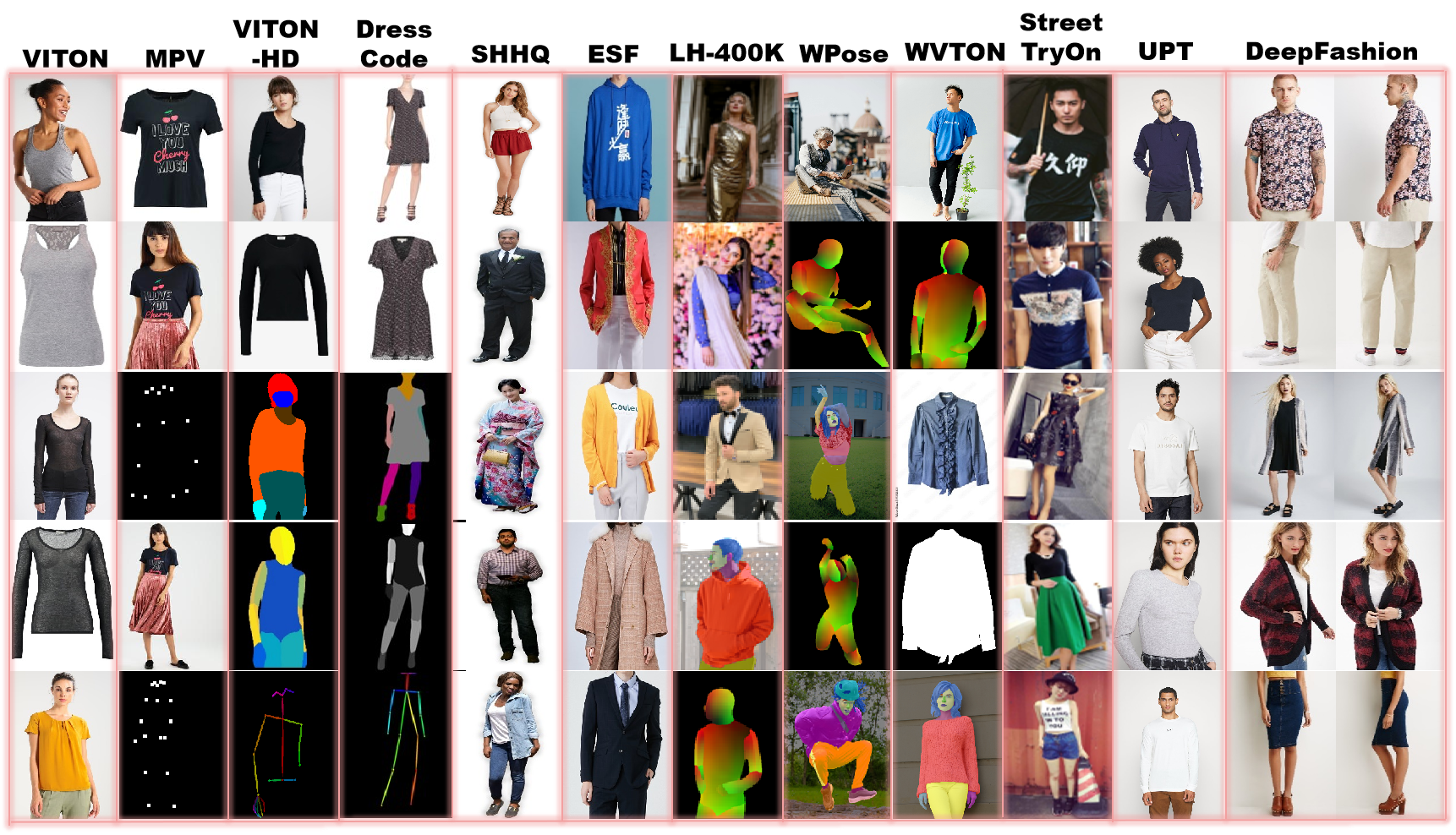}
\caption{
\reviseMark{Examples in existing datasets. }
}
\label{Dataset}
\end{figure*}

\section{Datasets and Evaluation Details}\label{Datasets and Evaluation Criteria}

\subsection{Datasets}\label{Datasets}
\reviseMark{The key properties of existing datasets for image-based virtual try-on are illustrated in Table \ref{table_dataset}, and Fig. \ref{Dataset} show some examples within these datasets. The datasets containing clothing images are commonly used for trying on in-shop clothes with paired annotation while those without clothing images usually demonstrate the performance of unpaired try-on. The property of multi-pose in Table \ref{table_dataset} does not refer to the pose types within the whole dataset but indicates whether one individual sample owns multiple poses. }

\reviseMark{\textbf{VITON} \citep{VITON} was the first collection for the task of trying on in-shop clothes and opens up a new research area, but currently are not publicly available due to ethical concerns.\textbf{ VITON-HD} \citep{VITON-HD} is an alternative dataset, with a resolution of $1024\times768$. It additionally contains parsing information such as segmentation, dense pose and pose keypoints. \textbf{Deep Fashion} \citep{Deepfashion} is a large-scale dataset including four subsets namely: Category and Attribute Prediction Benchmark, In-Shop Clothes Retrieval Benchmark, Consumer-to-shop Clothes Retrieval Benchmark, and Fashion Landmark Detection Benchmark. The In-Shop Clothes Retrieval Benchmark is often used for virtual try-on tasks, which contains product IDs and images of the same model in different poses. Each model in \textbf{MPV} \citep{MPV} corresponds to multiple images of different poses. \textbf{Dress Code} \citep{DressCode} contains upper-body, lower-body, and full-body clothing for both female and male, as well as corresponding model try-on images which are cut at the nose to ensure the privacy. }

\reviseMark{\textbf{SHHQ} \citep{SHHQ} aims at full-body human generation and has features of diverse attributes such as human poses, and characters' garment texture and type. \textbf{UPT} \citep{PASTA-GAN} contains person images in half-body and full-body frontal views, wearing a variety of garments. The clothed model images in \textbf{E-Shop Fashion (ESF)} dataset \citep{DGP_yuduiqi_ESF} are centrally aligned and filled to the region between the chin and thighs. }

\reviseMark{Recently, in-the-wild try-on attracts interests and the datasets where person images have diverse backgrounds are collected. \textbf{StreetTryOn} \citep{street} is a new benchmark for evaluating virtual try-on in natural environments with a wide range of clothing categories and human poses. \textbf{LH-400K, WPose and WVTON} are proposed by \cite{UniHuman} with a variety of backrgounds, ages and body sizes, respectively serving as training and testing datasets. Specifically, WPose focuses on diverse poses and WVTON concentrates on garments.}

\subsection{Evaluation Criteria}\label{Evaluation Criteria}
In this survey, methods are evaluated with previously used criteria such as SSIM (Structural Similarity Index Metric), IS (Inception Score), LPIPS (Learned Perceptual Image Patch Similarity) and FID (Frechet Inception Distance), and a new semantic score. These criteria cover the evaluation in aspects of individual or set comparison, structural or semantic similarity.

\textbf{SSIM:} Structural Similarity Index Metric  \citep{SSIM} is a traditional image similarity evaluation standard that is originally used for evaluating the performance of image compression algorithms. 
The inputs to compute SSIM are two individual images, i.e., the ground truth image and the generated image. SSIM value ranges within [0,1], where the larger value indicates the smaller difference between the compared two images and the better realism of the generated image. This criterion is sensitive to pixel shift, so high-level semantic feature is also considered for the evaluation.

\textbf{IS:} Inception Score  \citep{IS} is a metric based on Inception network, which is originally used to evaluate generation models in terms of clarity and diversity. 
The calculation of IS only involves generated images, and the prediction probability of generated images is used to describe the performance. 
The higher value of IS indicates the better generation performance. However, different from the original generation tasks, the generation target for virtual try-on is a fixed class such as human photos. It is not fair enough for this task (therefore not adopted in our unified evaluation), and FID becomes an alternative way which compares the distributions of generated images and ground truth.

\textbf{FID:} Frechet Inception Distance  \citep{FID} measures the statistical similarity between two sets of images by calculating the Frechet distance between the feature vectors of the real images and the generated images. Lower value of FID indicates smaller difference between two sets in feature space and better performance.

\textbf{LPIPS:} Like IS and FID, Learned Perceptual Image Patch Similarity  \citep{LPIPS} also utilizes a pretrained network (e.g., VGG  \citep{VGG} and AlexNet  \citep{Alexnet}) to capture high-level features. It calculates Euclidean or cosine distance between features output via multiple layers. Lower value of LPIPS indicates closer similarity of two images.

\begin{figure}[t!]
\centering
\includegraphics[width=2.9in]{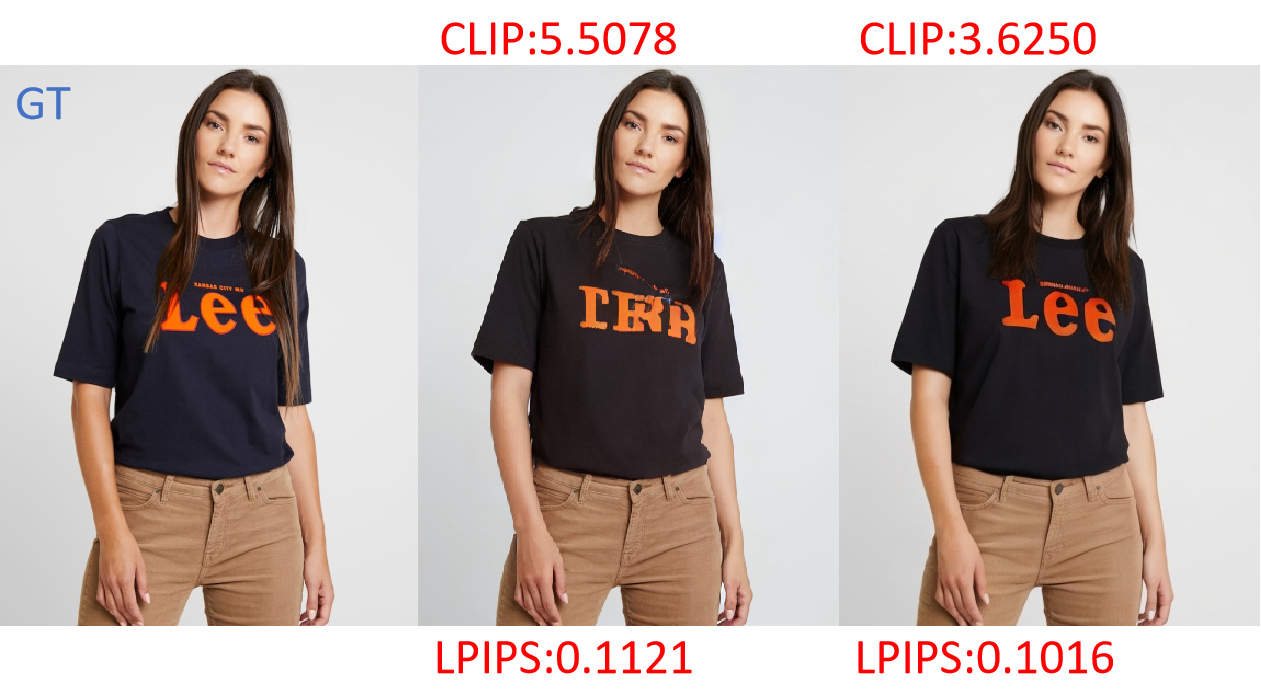}
\caption{\reviseMark{Comparison between semantic score and LPIPS. The scores for different generated results indicate that the proposed semantic score (denoted as CLIP in the figure) is more sensitive to semantic difference. }}
\label{clip_advantage}
\end{figure}

\textbf{Semantic Score:} 
\reviseMark{We propose a new semantic score with the help of recent advanced model CLIP  \citep{clip} for evaluating image-based virtual try-on. The Euclidean distance between CLIP features of generated image and ground truth is calculated, so lower value indicate closer similarity in semantic information. Compared with previous pretrained network such as VGG and AlexNet used in IS/FID/LPIPS, CLIP is suited for this task due to its training on a diverse range of images and text, enabling it to capture a broad spectrum of visual and semantic features. Unlike traditional metrics like SSIM that primarily focus on pixel-level accuracy, CLIP assesses the semantic coherence between the generated and ground truth images, which is critical in evaluating how realistically the garments are portrayed in virtual try-on. Fig. \ref{clip_advantage} also shows that the semantic score has advantages in evaluating the semantic difference for virtual try-on. The scores above and below the subfigure are computed by comparing the generated result with ground truth (i.e., the left-most subfigure). Although both LPIPS and CLIP indicate that the right result is better, CLIP exhibit a large gap between the two results.}

\reviseMark{It is worth noting that we for the first time separately evaluate the clothing warping performance and generation ability for non-try-on area with a human parser \citep{Dresscode_seg}. 
The generation quality of try-on area and non-try-on area reflects the fidelity of clothing warping and the preservation of human characteristics, which could provide existing methods valuable information for further improvement. }

\subsection{\reviseMark{Implementation Details}}
\label{sec:implementation_details}
\reviseMark{
Due to the ethical concern raised by VITON dataset, we use VITON-HD dataset for the unified evaluation. 
Re-training previous methods cannot guarantee the optimal parameters for fair comparison, so we use officially pretrained weights from each respective method for inference. Although the two datasets come from the same source which indicate their distributions are similar, we still mark the methods which are trained on VITON dataset while tested on VITON-HD dataset as cross-dataset evaluation. }

\reviseMark{The resolution of generated images follows the implementation of officially released code. Specifically, the resolution of VITON, CP-VTON, LM-VTON, CP-VTON+, PFAFN, DCTON, Style-Flow and SDAFN is $256\times192$. The resolution of VITON-HD and HR-VITON is $1024\times768$. The resolution of Ladi-VTON, DCI-VTON, GP-VTON, CAT-DM, Stable VITON and TPD is $512\times384$. As the evaluation criteria are affected by image resolution, the methods with the same resolution are put together for comparison. CLIP \citep{clip} could take images with any resolution as input and resize the resolution to $224\times224$ for later computation. }


\begin{figure}[t!]
\centering
\includegraphics[width=3.0in]{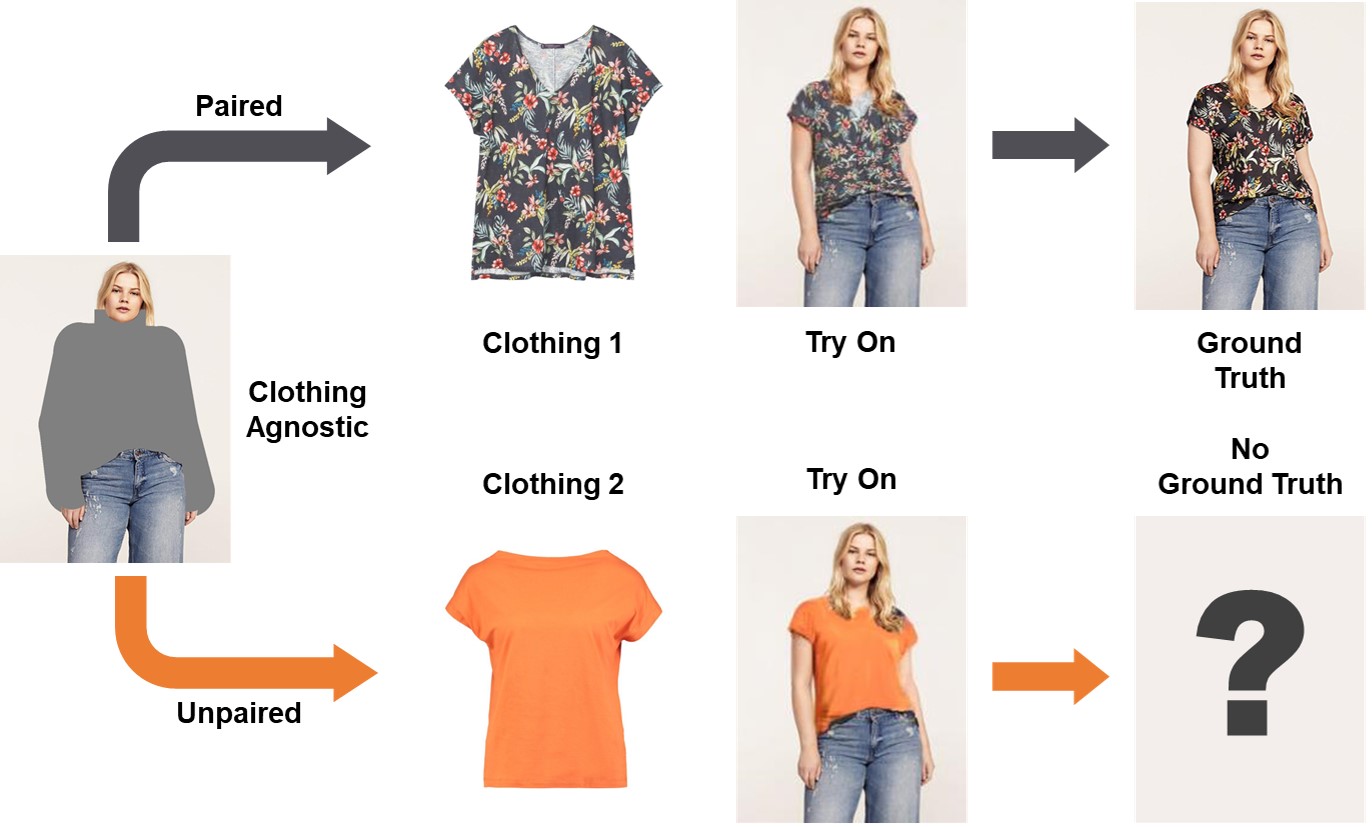}
\caption{\reviseMark{Two testing conditions are involved for evaluation. Paired evaluation refers to the condition that the clothing image and person image are paired and the clothing-agnostic image is parsed from the person image. Unpaired evaluation refers to that the clothes in the clothing image and person image are different, so there is no ground truth for the try-on result.}}
\label{paired-unpaired}
\end{figure}

\section{Experimental Results and Analysis}\label{Experimental Results and Analysis}

\label{sec:experiments}
Due to the lack of unified evaluation and comprehensive analysis, we run existing open-source codes in image-based virtual try-on area and evaluate them with the same referees.
In this section, experimental results are given as quantitative results, qualitative results and user study. Besides the observations towards these results, suggestions in terms of method design are also analyzed. 

\subsection{Quantitative Results}\label{Quantitative Results}
\reviseMark{Previous methods are quantitatively evaluated in terms of testing conditions and areas. 
Fig. \ref{paired-unpaired} shows two testing conditions, i.e., paired and unpaired. Paired situation refers to that we have a pair of clothing image and image of person wearing the same clothes. Such data are abundant on the Internet and usually used for training by masking the person image to obtain clothing agnostic person representation.
Unpaired situation is close to practical scenario where the clothing in the person image is different from target clothing, where we do not have the ground truth for try-on image. }

\begin{figure}[!t]
\centering
\includegraphics[width=2.7in]{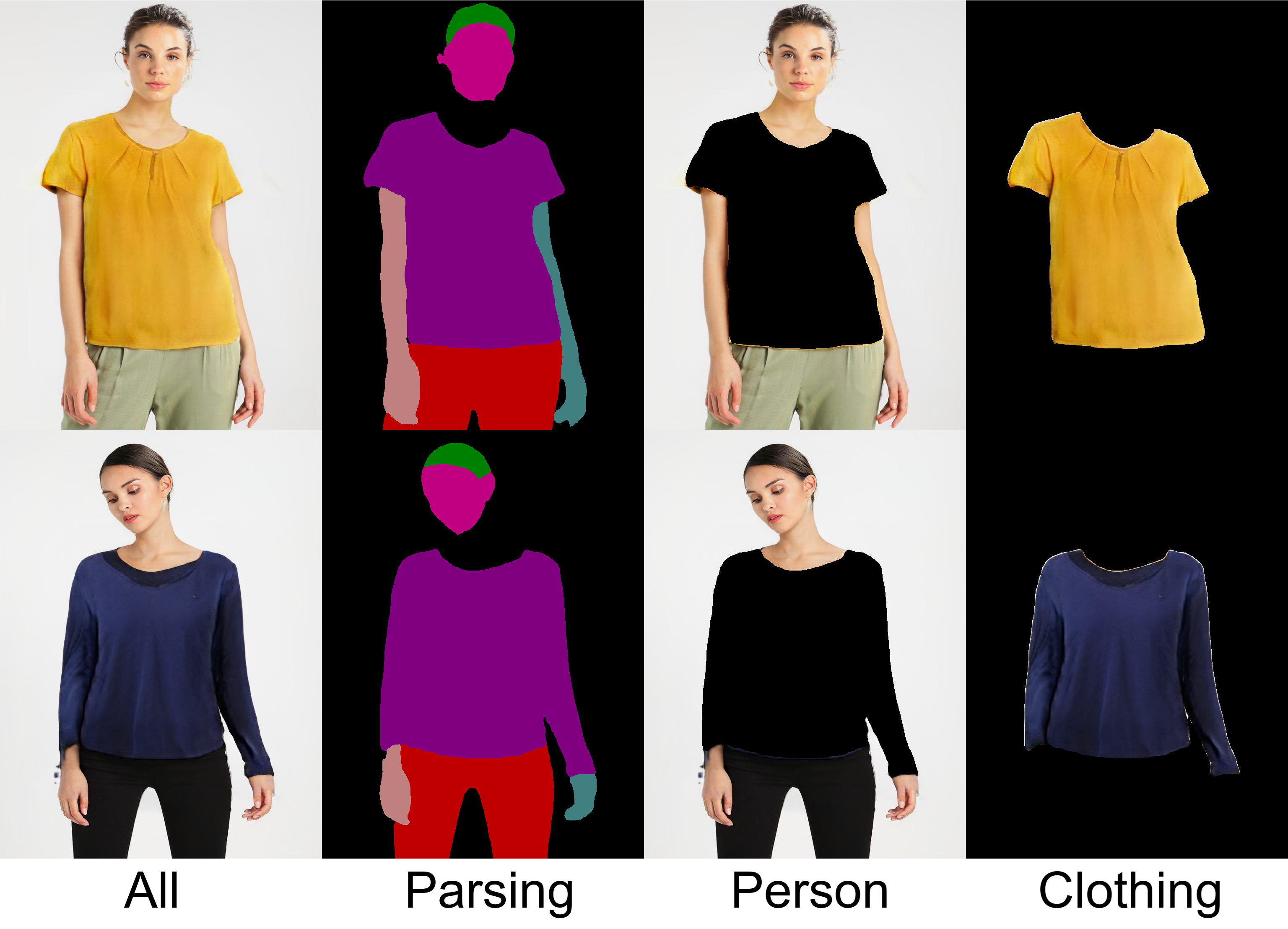}
\caption{\reviseMark{Three testing conditions (``All", ``Person" and ``Clothing"). With the help of a parser \citep{Dresscode_seg}, we separate the generated image to non-try-on area and clothing area. The deleted areas are replaced with black pixel.}}
\label{evaluation_area}
\end{figure}

\reviseMark{We for the first time separately evaluate the generation quality of different areas for virtual try-on, i.e., the warped clothing area, the non-try-on area, and the whole area, which are respectively denoted as ``Clothing", ``Person" and ``All". Fig. \ref{evaluation_area} shows several examples for different evaluation areas. The parsing labels of ground truth via a recent parser \citep{Dresscode_seg} are used to extract specific area for comparison under paired situation. For unpaired case, we should separately parse the ground truth and generated results for comparison. It should be noted that the parsers in training and testing for the comparison methods keep the same as the official implementation, which might be different from the parser used in the separated evaluation. And the parser for evaluation could be changed with the advancement of human parsing.}

\begin{figure*}[!t]
	\centering
        \subfloat[SSIM \textbf{$\uparrow$}]{
		\includegraphics[width=0.99\linewidth]{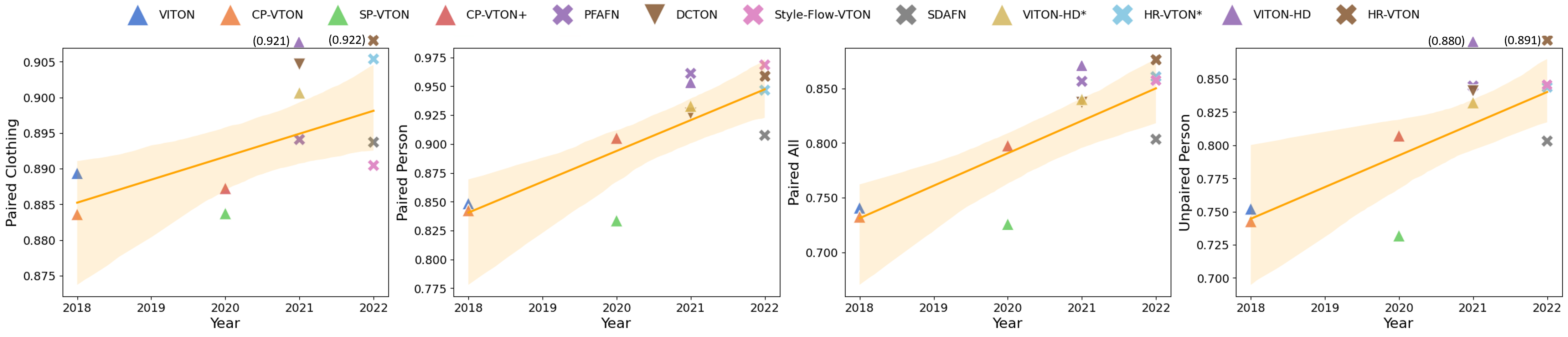}
            \label{SSIM_score}}

        \centering
	\subfloat[FID \textbf{$\downarrow$}]{
		\includegraphics[width=0.99\linewidth]{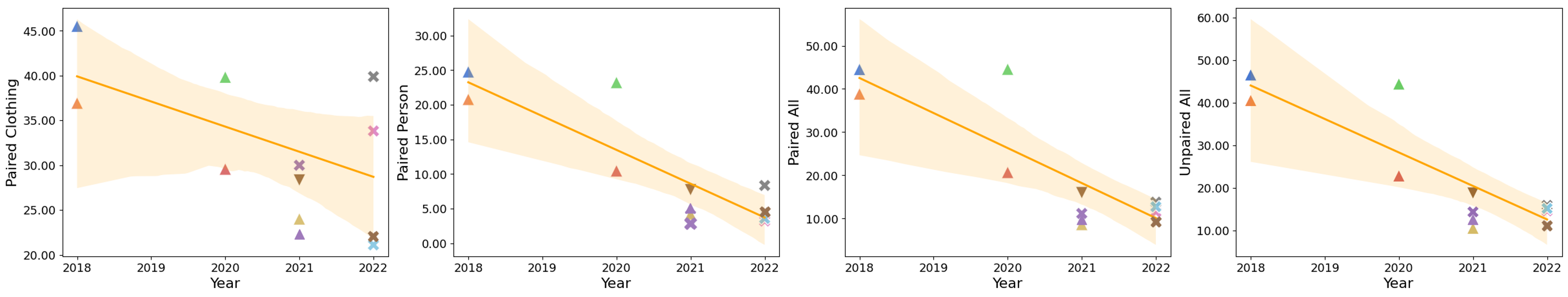}
            \label{FID_score}} 
        \centering

        \centering
	\subfloat[LPIPS \textbf{$\downarrow$}]{
		\includegraphics[width=0.99\linewidth]{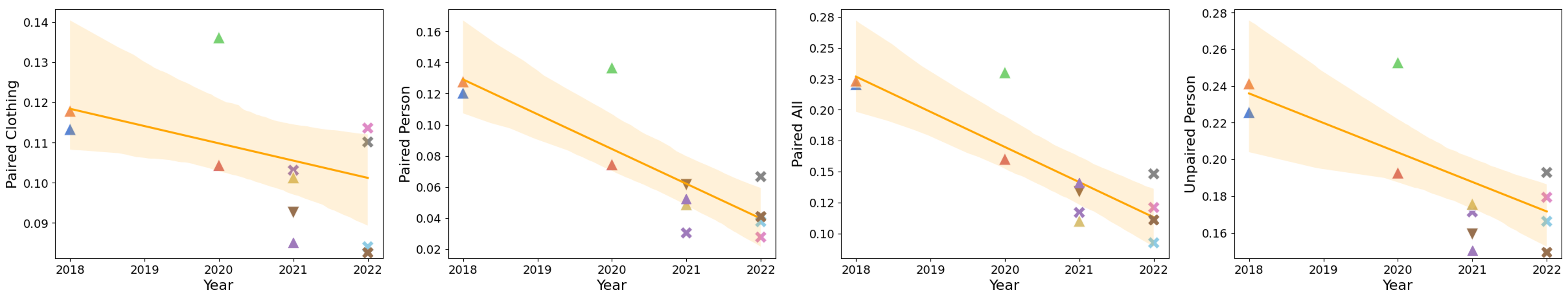}
            \label{LPIPS_score}} 

        \centering
	\subfloat[Semantic Score \textbf{$\downarrow$}]{
		\includegraphics[width=0.99\linewidth]{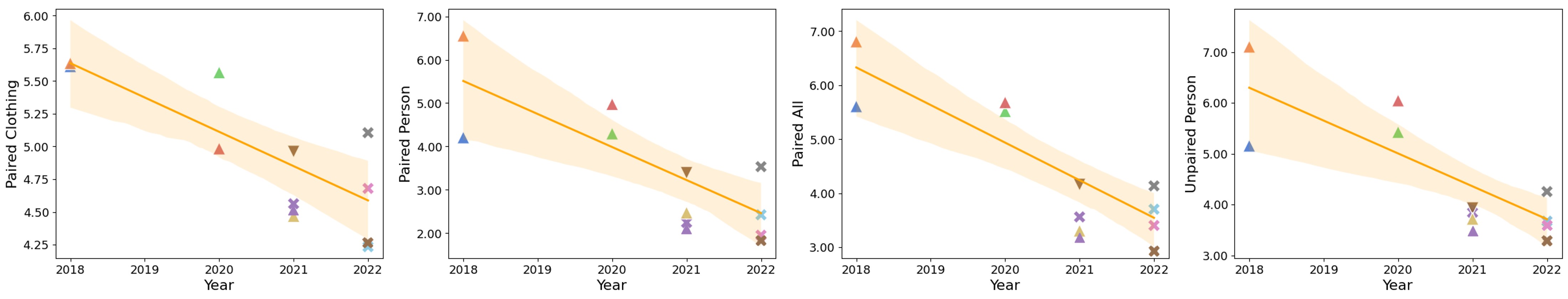}
            \label{Clip_score}}

        \caption{\reviseMark{Cross-dataset quantitative results on VITON-HD. The vertical axis indicates the testing condition is paired or unpaired and the evaluation target is clothing, non-try-on area (denoted as ``person") or all together. For method symbol, the cross symbol indicates the clothing warping strategy is flow while the triangle symbol represents TPS-based method. VITON-HD* and HR-VITON* denote that we resize the image resolution from $1024\times768$ to $256\times192$ via the default resize function in python imaging library. }
        }
    \label{fig:quantitative}
\end{figure*}

\begin{figure*}[!t]
	\centering
        \subfloat[SSIM \textbf{$\uparrow$}]{
		\includegraphics[width=0.99\linewidth]{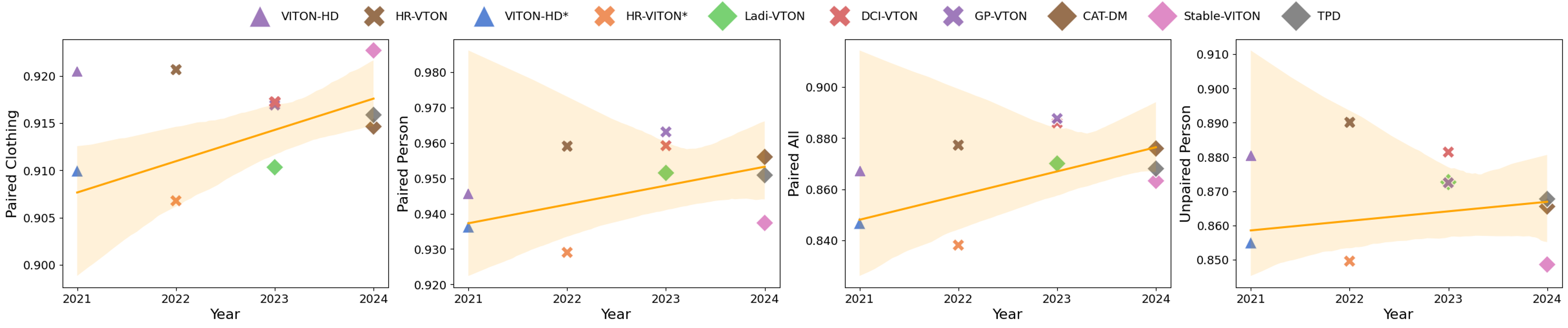}
            \label{SSIM_score}}

        \centering
	\subfloat[FID \textbf{$\downarrow$}]{
		\includegraphics[width=0.99\linewidth]{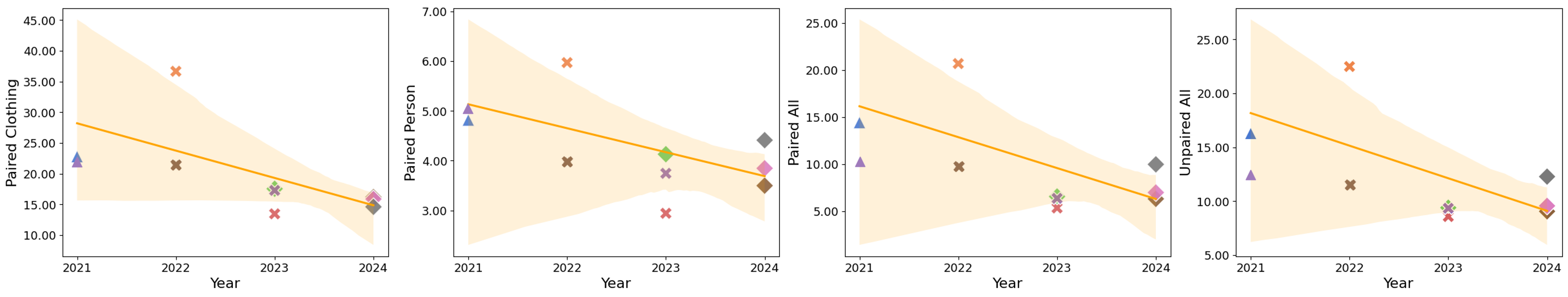}
            \label{FID_score}} 
        \centering

        \centering
	\subfloat[LPIPS \textbf{$\downarrow$}]{
		\includegraphics[width=0.99\linewidth]{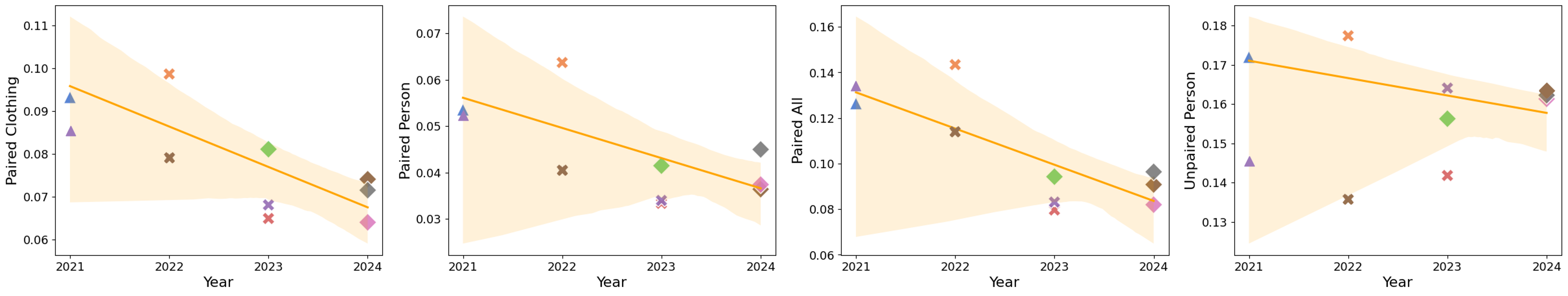}
            \label{LPIPS_score}} 

        \centering
	\subfloat[Semantic Score \textbf{$\downarrow$}]{
		\includegraphics[width=0.99\linewidth]{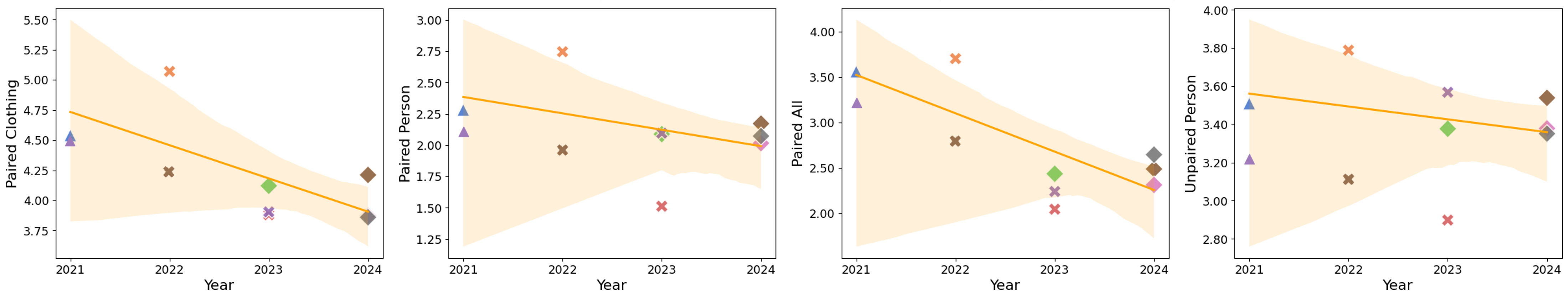}
            \label{Clip_score}}

        \caption{\reviseMark{Quantitative results on VITON-HD. VITON-HD* and HR-VITON* denote that we resize the image resolution  from $1024\times768$ to $512\times384$ via the default resize function in python imaging library.}}
    \label{fig:quantitative2}
\end{figure*}

The quantitative results of representative methods in the order of publication year are shown in Fig. \ref{fig:quantitative} and  \ref{fig:quantitative2}. As unpaired condition lacks ground truth, it is hard to evaluate the generation quality of full body for SSIM, LPIPS and Semantic Score. FID could be computed because it measures the similarity of two image sets. 
\reviseMark{It should be noted that the evaluation criteria are sensitive to image resolution, so the quantitative results are compared under the same resolution. The image resolution of the comparison methods has been illustrated in Sec. \ref{sec:implementation_details}. To save space, we did not draw separate figure for VITON-HD \citep{VITON-HD} and HR-VITON \citep{HR-VTON}, which are under the resolution of $1024\times768$. We put their results to both Fig. \ref{fig:quantitative} and  \ref{fig:quantitative2}. }

From the quantitative results, we have the following observations:


\begin{figure*}[t!]
\centering
\includegraphics[width=6.3in]{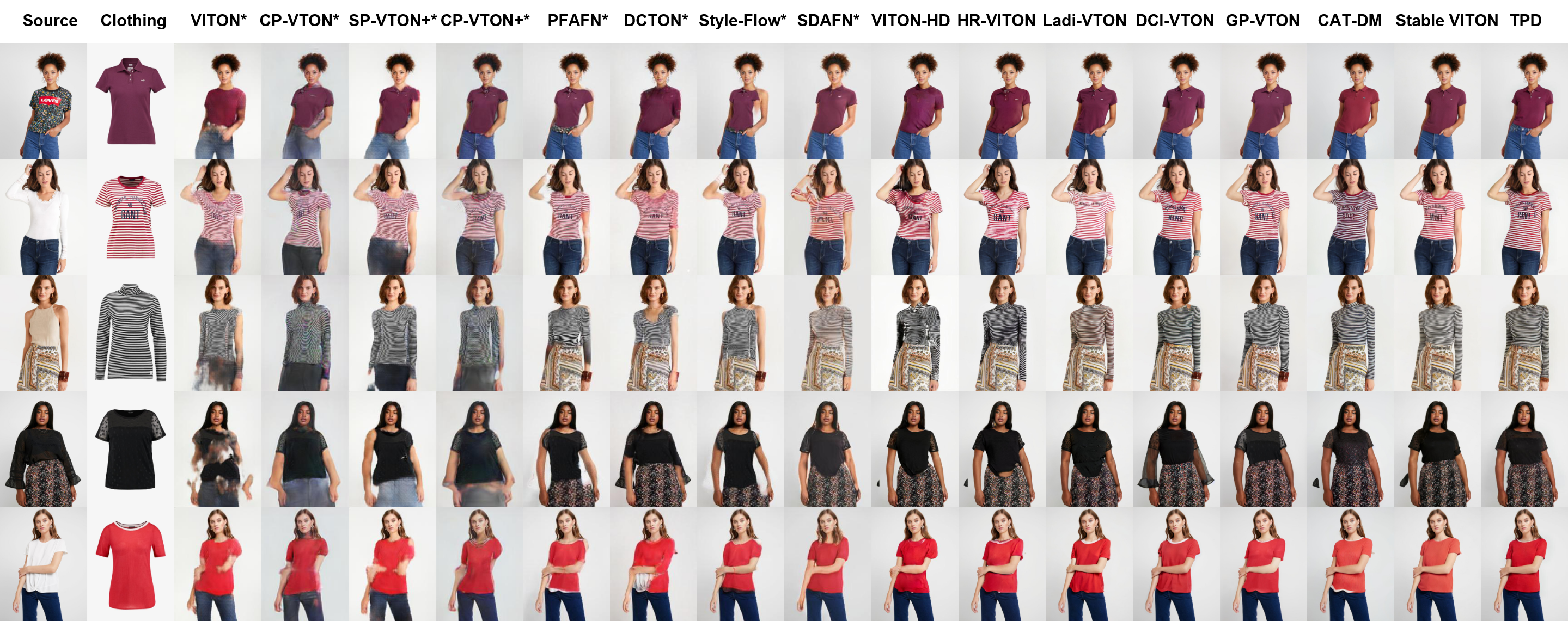}
\caption{\reviseMark{Qualitative results on VITON-HD, with variations of person poses, shapes and clothing patterns. The methods marked with * use the officially pretrained weights trained on VITON for inference.}}
\label{keshihua_all}
\end{figure*}

\begin{itemize}
    \item On all criteria, the results under paired condition surpass those under unpaired condition. It is understandable that paired testing is a easier task than unpaired situation, as methods are usually trained with paired data.
    \item By comparing the results of ``Paired Clothing" with ``Paired Person", the former falls behind the latter due to the challenges of clothing warping.
    \item The comparative performance of representative methods on these criteria are roughly consistent, which to some extent show the consistency of evaluation metrics.
    \item \reviseMark{For high-resolution methods, HR-VTON  
    \citep{HR-VTON} performs better in Fig \ref{fig:quantitative}}, which attributes to its dual-path pipeline that simultaneously predict human body and clothing and the deformed clothing is further aligned with the predicted semantic segmentation.
    \item With VITON-HD and HR-VITON as a bridge, by comparing Fig. \ref{fig:quantitative} and \ref{fig:quantitative2} we can find that the overall generation quality of diffusion-based methods outperforms GAN-based methods.
\end{itemize}

\subsection{Qualitative Results}\label{Qualitative Results}
\reviseMark{The qualitative results are displayed in Fig. \ref{keshihua_all}} by showing several cases of image-based virtual try-on. The $1^{st}$ row the figure shows an easy task where these representative methods are good at, and we can find that recent methods generate results with high-fidelity and fine-grained details. 
The preservation of clothing texture such as special patterns and stripes (e.g., the $2^{nd}$, $3^{rd}$ and $4^{th}$ rows) could be further improved. GAN-based methods usually generate blurry results while diffusion-based methods face challenges in controllability. 
Person shape (e.g., $4^{th}$ row) has less effects while challenging pose (e.g., the last row) will increase the difficulty of try-on. Challenging cases will be further discussed in Sec. \ref{sec:unresolved}.

\reviseMark{For earlier works such as VITON  \citep{VITON}, SP-VTON  \citep{sp-VITON} and CP-VTON  \citep{CP-VTON} models, the focus is on the processing of the try-on part, which roughly align the clothes to body. The quality of the results generated by PFAFN  \citep{PFAFN}, Style-Flow-VTON  \citep{style-flow} and SDAFN  \citep{SDFN} gets better, where the boundary between the body area and the clothing area is more natural and clear. It mainly attributes to the warping ability of flow. VITON-HD  \citep{VITON-HD} and HR-VITON  \citep{HR-VTON} are two methods designed for high-resolution virtual try-on, and the results of HR-VITON contain more details. Overall, the results of diffusion-based methods (in Fig \ref{keshihua_all}) show clear appearance. GP-VTON \citep{GP-VTON} is a GAN-based method, but shows competitive performance, which attributes to its clothing warping strategy of separately warping partitions. DCI-VTON \citep{DCI-VTON} first warps clothing and then use diffusion model to refine the try-on results, which both controls the fidelity of original clothes and shows clear appearance. Although diffusion-based methods (e.g., Ladi-VTON  \citep{LaDI-VTON}, CAT-DM \citep{CAT-DM}, Stable VITON \citep{Stable-VITON} and TPD \citep{TPD}) show clear appearance, the controllability in clothing color and texture can still be improved.}


\begin{figure*}[t!]
\centering
\includegraphics[width=6.3in]{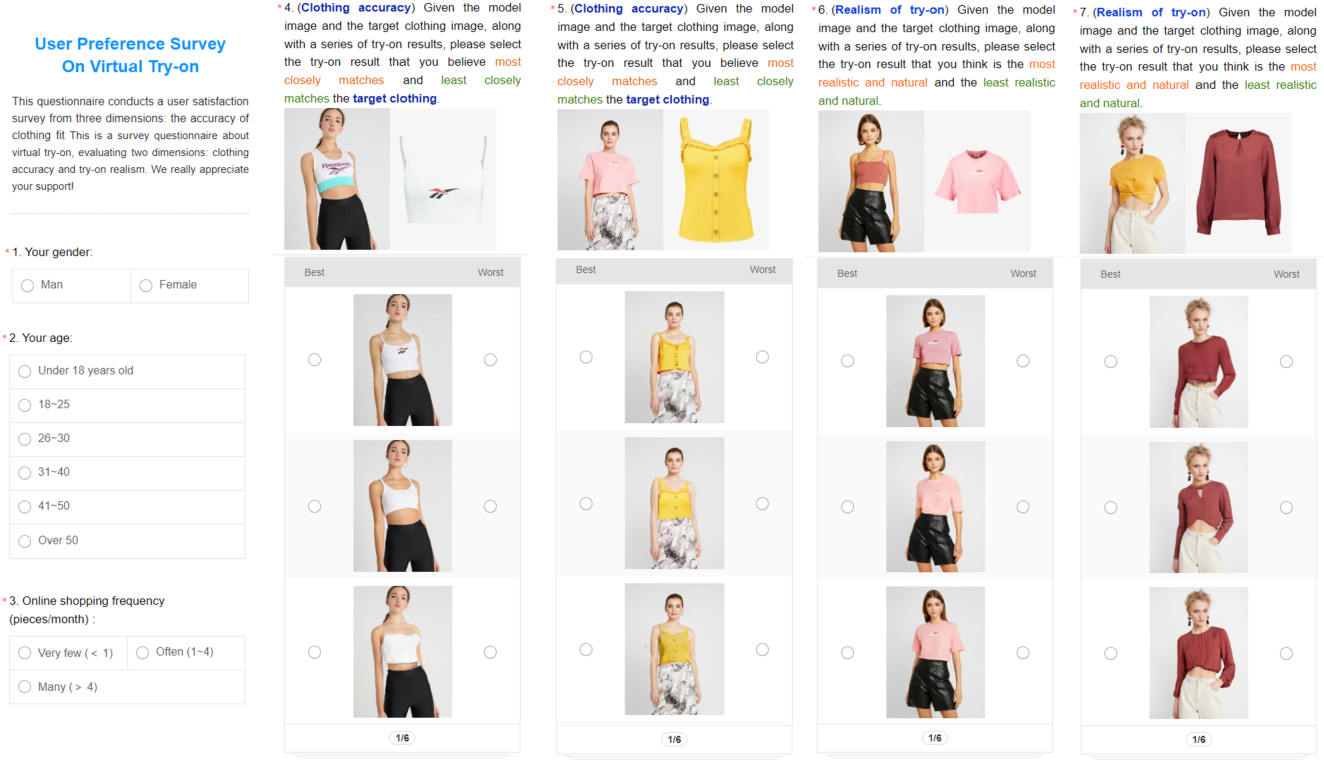}
\caption{\reviseMark{The interface of user study.}}
\label{fig:interface}
\end{figure*}

\begin{figure*}[htbp] 
    \centering
    \begin{subfigure}{0.32\textwidth} 
        \centering
        \includegraphics[width=0.98\textwidth]{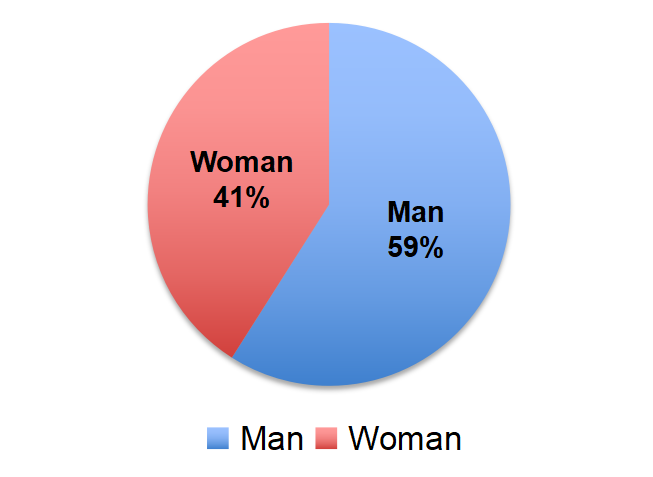}
        \caption{\reviseMark{Gender}}
        \label{fig:user_gender}
    \end{subfigure}
    \hfill 
    \begin{subfigure}{0.32\textwidth}
        \centering
        \includegraphics[width=0.98\textwidth]{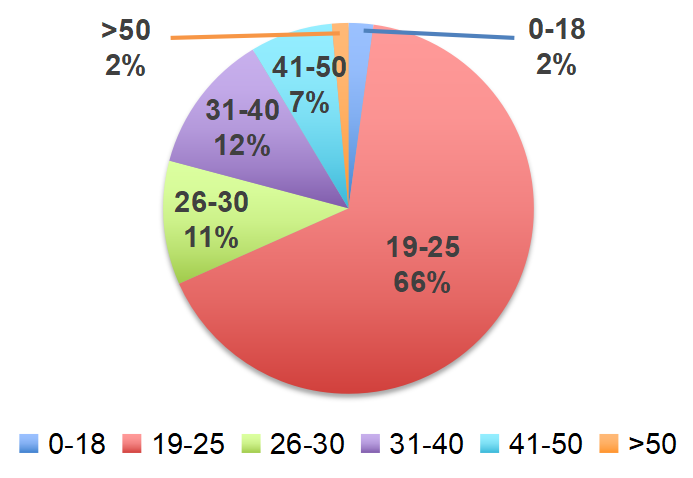}
        \caption{\reviseMark{Age}}
        \label{fig:user_age}
    \end{subfigure}
    \hfill
    \begin{subfigure}{0.32\textwidth}
        \centering
        \includegraphics[width=0.98\textwidth]{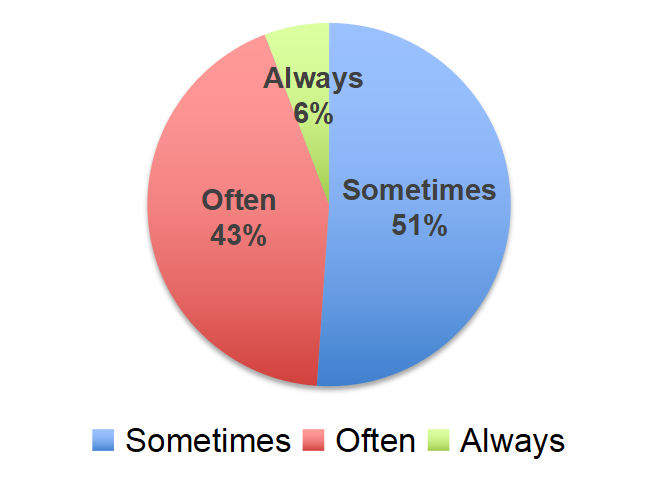}
        \caption{\reviseMark{Frequency}}
        \label{fig:online_frequency}
    \end{subfigure}
    \caption{\reviseMark{Demography of participants: (a) gender, (b) age and (c) frequency of shopping online.} }
    \label{fig:user_study_demo}
\end{figure*}

\subsection{User Study}\label{User Study}
\reviseMark{In order to assess the visual quality of virtual try-on results in human perspective, we collect public questionnaires online. The questionnaire is designed to evaluate user preference towards 16 comparison methods in aspects of clothing accuracy and fitting realism. Clothing accuracy aims to reflect the quality of clothing reconstruction and fitting realism depends on the alignment bewteen garment and human body. }

\reviseMark{Different from straightforwardly rating 16 results, we choose MaxDiff (Maximum Difference Scaling, also known as Best-Worst Scaling) for our evaluation. We choose MaxDiff for the following main reasons: Firstly, humans are very picky about virtual try-on results, which may give similar low ratings to existing methods (leading to bias or skewed results). MaxDiff forces participants to make distinct choices, which reduces the likelihood of neutral or invalid responses and provides clearer insights into preferences.
Secondly, choosing the most and least preferred items is more intuitive and straightforward than rating each method on a scale, making the process less cognitively demanding and the result objective.}

\begin{figure}[!t]
\centering
\includegraphics[width=3.0in]{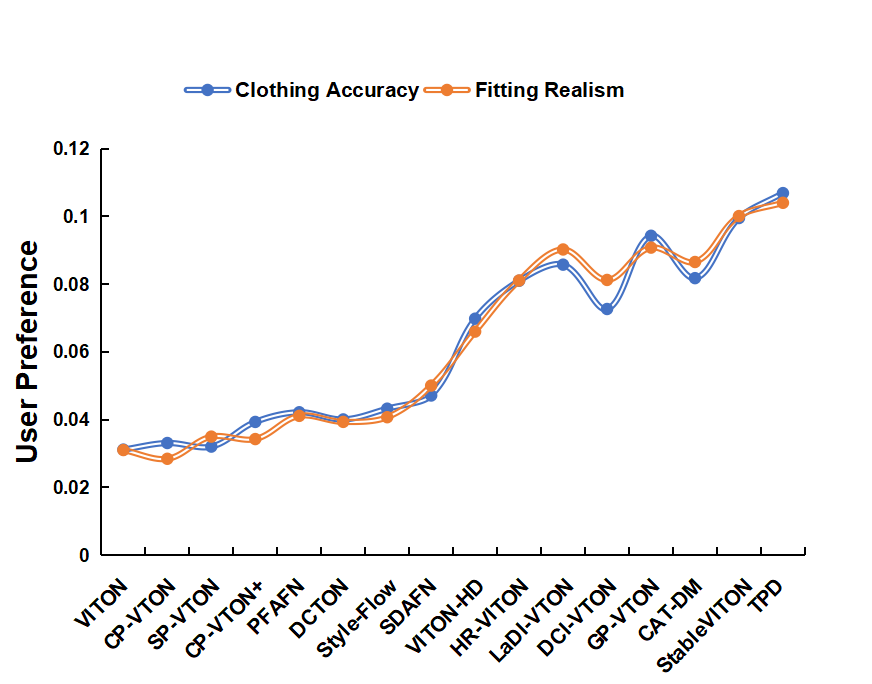}
\caption{\reviseMark{User preference towards existing methods.}}
\label{fig:user_preference}
\end{figure}

\reviseMark{As shown in the interface figure (Fig. \ref{fig:interface}), to ensure that participants have enough patience to answer, each questionnaire contains 4 samples and for each sample we set up 6 best-worst questions with 3 options. The 4 samples (2 for clothing accuracy and 2 for fitting realism) in each questionnaire are randomly selected, and different questionnaires are probably different due to the large pool and random setting.}


\begin{figure*}[t!]
\centering
\includegraphics[width=6.3in]{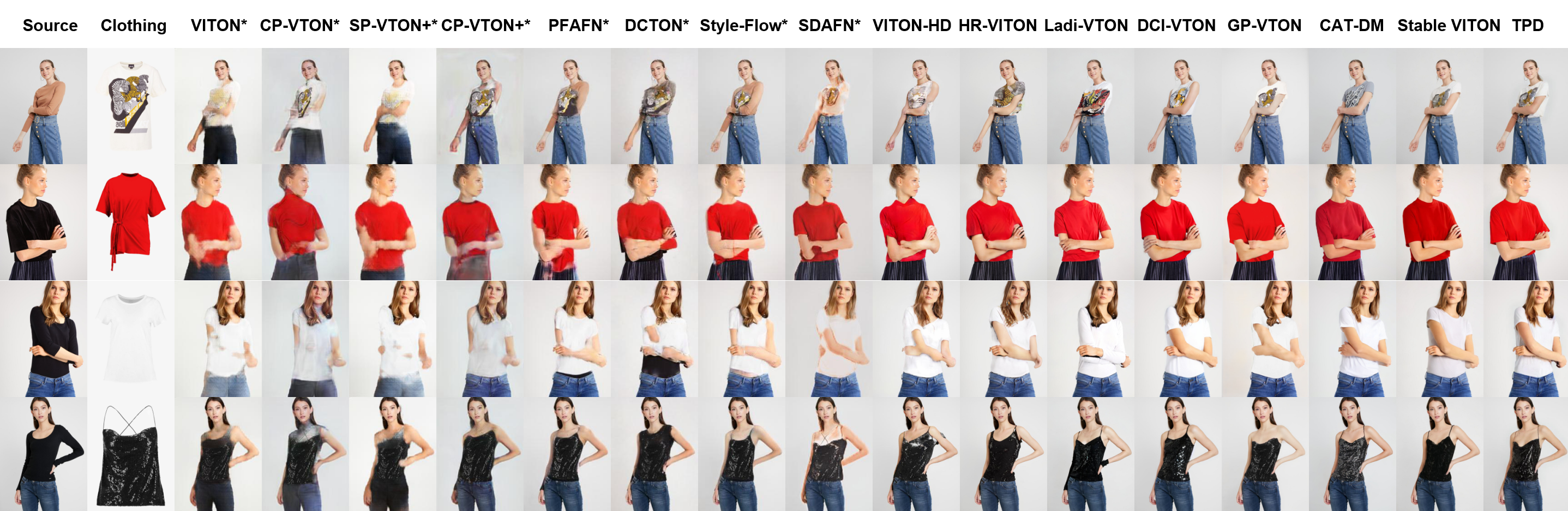}
\caption{\reviseMark{Performance under challenging poses on VITON-HD. The methods marked with * use the officially pretrained weights trained on VITON for inference.}}
\label{fig:challenging_poses2}
\end{figure*}

\reviseMark{We received a total of $139$ valid questionnaires from anonymous participants, based on which we compute the demography of participants (Fig. \ref{fig:user_study_demo}) and user preference (Fig. \ref{fig:user_preference}). The score for user preference is calculated via MaxDiff based on multinomial logit model. From Fig. \ref{fig:user_preference} we have the following two observations: 1) Current methods show similar performance in clothing accuracy and fitting realism; and 2) Earlier GAN-based methods could hardly satisfy users for virtual try-on and recent methods improve a lot, which can also be validated in the visual results.}


\subsection{Discussion on Method Design}\label{Discussion on Method Design}
By analyzing the experimental results of existing methods, we have the following observations:

\begin{itemize}
    \item \textbf{Pipeline perspective.} The structures of existing pipelines are classified as shown in Fig. \ref{totalmodel}, and the specific pipeline adopted by each approach is illustrated in Table \ref{big_table}. It could be found that there is no obvious pipeline preference in the development of image-based virtual try-on methods, and recent methods tend to design one-stage pipeline (i.e., type \Rmnum{1}) with diffusion models. Except recent diffusion-based methods, the dual-path pipeline (i.e., type \Rmnum{7}) achieves superior performance where the dual paths facilitate each other to optimize the generation performance.
    \item \textbf{Clothing warping perspective.} Existing warping approaches adopted by representative methods are shown in Table \ref{big_table}. As illustrated in Sec. \ref{Cloth Warping}, spatial transformation network (STN) usually cooperates well with thin plate spline (TPS) transformation, and TPS plays a dominating role in the development of clothing warping methods. The performance of clothing warping approaches is reflected as the results of ``paired clothing" in Fig. \ref{fig:quantitative}, where the triangle symbol represents TPS-based method and the cross mark denotes flow-based method. Comparatively, the warping method of flow estimation shows superior performance with more flexible transformation. Additionally, Table \ref{big_table} shows that most diffusion-based methods do not contain an explicit clothing warping module, indicating that this module becomes not necessary with the improved performance of generation network.
    \item \textbf{Try-on perspective.} The final try-on performance depends on all procedures including try-on indication, cloth warping and try-on. The generation quality gets better with the development of generative model such as StyleGAN and Diffusion model.
\end{itemize}

\section{Unresolved Issues}\label{Unresolved Issues}
\label{sec:unresolved}
As illustrated in Table \ref{big_table}, most methods rely on human parsing. Human parsing plays a vital role in image-based virtual try-on, but also causes the main issues due to imperfect parsing ability. In this section, we further show examples for issues caused by human parsing. 

\subsection{Challenging Poses}\label{sec:challenging_poses}

Challenging pose usually attributes to the occlusion caused by arms. Take the arm-crossed posture (Fig. \ref{fig:challenging_poses2}) for example, three problems should be overcome to perform well: 
a. correctly judge the position of the arm in front or back of the body and generate the right mask to put on clothes; 
b. correctly generate the arms and hands in a crossed state; 
and c. correctly handle the relationship between the clothing on the arm and the exposed skin.

\begin{figure}[!t]
\centering
\includegraphics[width=2.5in]{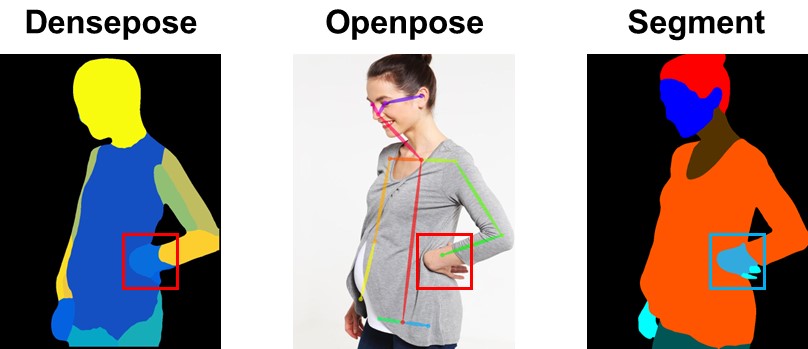}
\caption{Coarse pose parsing. Left: human parsing by densepose  \citep{Densepose}; Middle: pose keypoints estimated by openpose  \citep{Pose_VITON-HD(Openpose)}; Right: semantic parsing via  \citep{Seg_VITON-HD}. Current parser lacks fine-grained representation, e.g., the state of fingers at bent wrist.}
\label{chayao}
\end{figure}

\reviseMark{For problem a, from Fig. \ref{fig:challenging_poses2} we can find that most methods can tell the front or back position correctly except the ambiguous clothing with plain color.
For problem b, the crossed arms are not well solved. HR-VTON \citep{HR-VTON} and VITON-HD \citep{VITON-HD} mitigate this problem by optimizing the misaligned parts of the human body and clothing within the generator. 
The arms-crossed pose contains the crossing of the fingers and the arm, which requires high granularity generation ability of the model. Take Fig. \ref{chayao} for example, current representation such as Openpose  \citep{Pose_VITON-HD(Openpose)}, Densepose  \citep{Densepose} and semantic segmentation map  \citep{Seg_VITON-HD, Dresscode_seg} cannot label the fine-grained pose at the wrist, which fundamentally leads to the poor generation quality. 
Recent diffusion-based methods show high realism at fine-grained appearance.
The relationship between clothing and body is also well addressed with the development of methods.
However, it should be pointed out that current diffusion models are not fully controllable in terms of the generated contents.}

\begin{figure}[!t]
\centering
\includegraphics[width=3in]{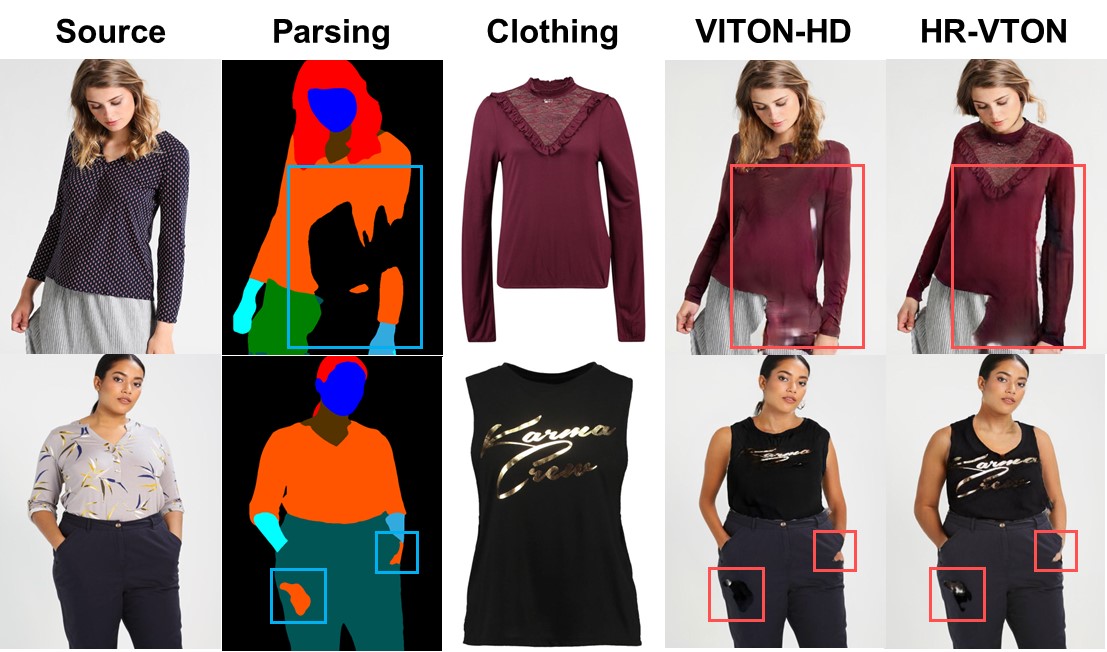}
\caption{Artifacts caused by noisy parsing.}
\label{yuyicuowu}
\end{figure}

\subsection{Limited Human Parsing}

\begin{figure*}[t!]
\centering
\includegraphics[width=6.3in]{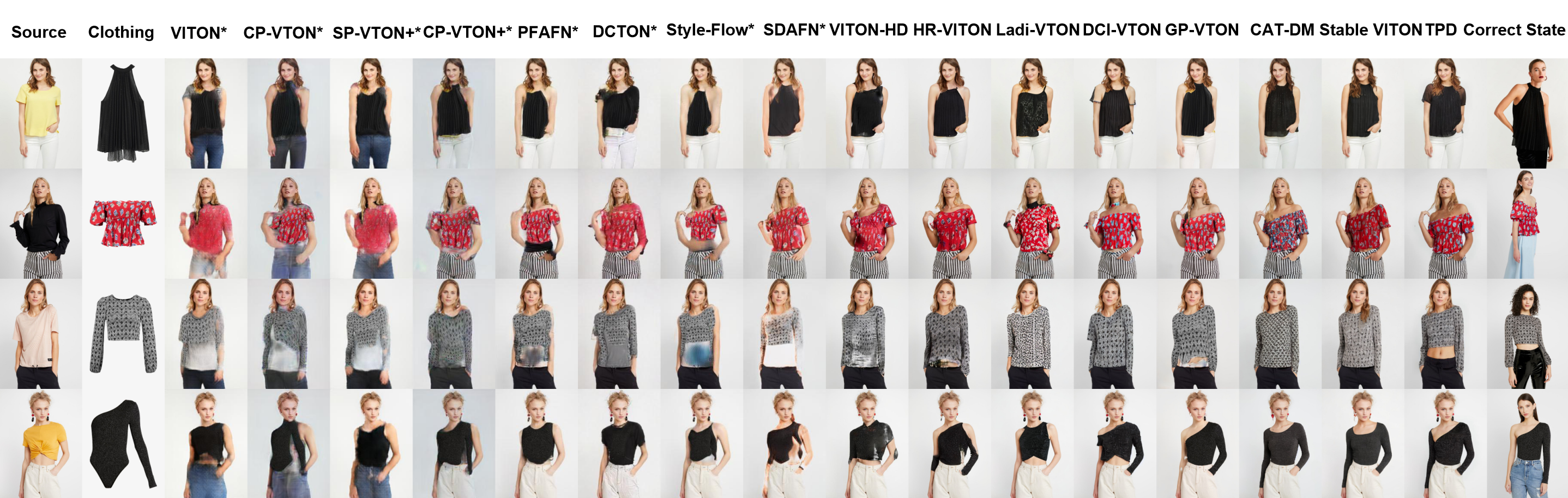}
\caption{\reviseMark{Changed clothing styles on VITON-HD. The right-most figures show the correct state of target clothing. The methods marked with * use the officially pretrained weights trained on VITON for inference.}}
\label{chuanyipianjian2}
\end{figure*}

Besides the pose estimation, image-based virtual try-on methods also rely on some other pre-processing results (as aforementioned in Sec. \ref{Cloth-agnostic Person Representation}). Fig. \ref{yuyicuowu} shows some results generated by HR-VITON and VITON-HD using noisy semantic information, where the wrong parser area has artifacts. It could be found in Fig. \ref{fig:person_representation} that $\mathcal{P}_{\substack{6,7,10,11}}$ involve information of original clothes expected to take off. For methods which adopt such representation (e.g., VITON  \citep{VITON}, CP-VTON  \citep{CP-VTON} and LM-VTON  \citep{LM-VITON}), the try-on results are affected by the style of original clothes. Take an example for these methods, the collar style of target clothes changes to that of the original clothes as shown in Fig. \ref{chuanyipianjian2}. 

Clothing style is an important aspect to show the performance of virtual try-on. \reviseMark{Previous efforts have been made to adjust the style or size of garments via association with landmarks or skeleton \citep{size_dose_matter,SAL-VTON,RP-VTON}. However, as shown in Fig. \ref{chuanyipianjian2}, it is difficult to keep the original style when the target clothing has unconventional styles.}
Compared with the correct wearing state of the corresponding clothing in the rightmost column of each row, the try-on results generated by the model have obvious biases, and the model tends to make the clothing fit perfectly on the upper body of the person body, presenting a slimming effect. 
This is unfair for clothing with loose styles or special designs in length, such as the results in the second and third rows of Fig. \ref{chuanyipianjian2}, where the short T-shirts are forcibly stretched to the normal length of T-shirts, and the long T-shirts are forcibly reduced/truncated to the normal length of T-shirts.

Essentially, the model does not really understand the clothing to be tried on, and it only deforms the clothing image and fills it in the original clothing area for replacement. \reviseMark{The reliance on the parsing of original clothes also introduces bias to inpainting area.}

\subsection{Limited Clothing Parsing}

\begin{figure}[t!]
\centering
\includegraphics[width=3in]{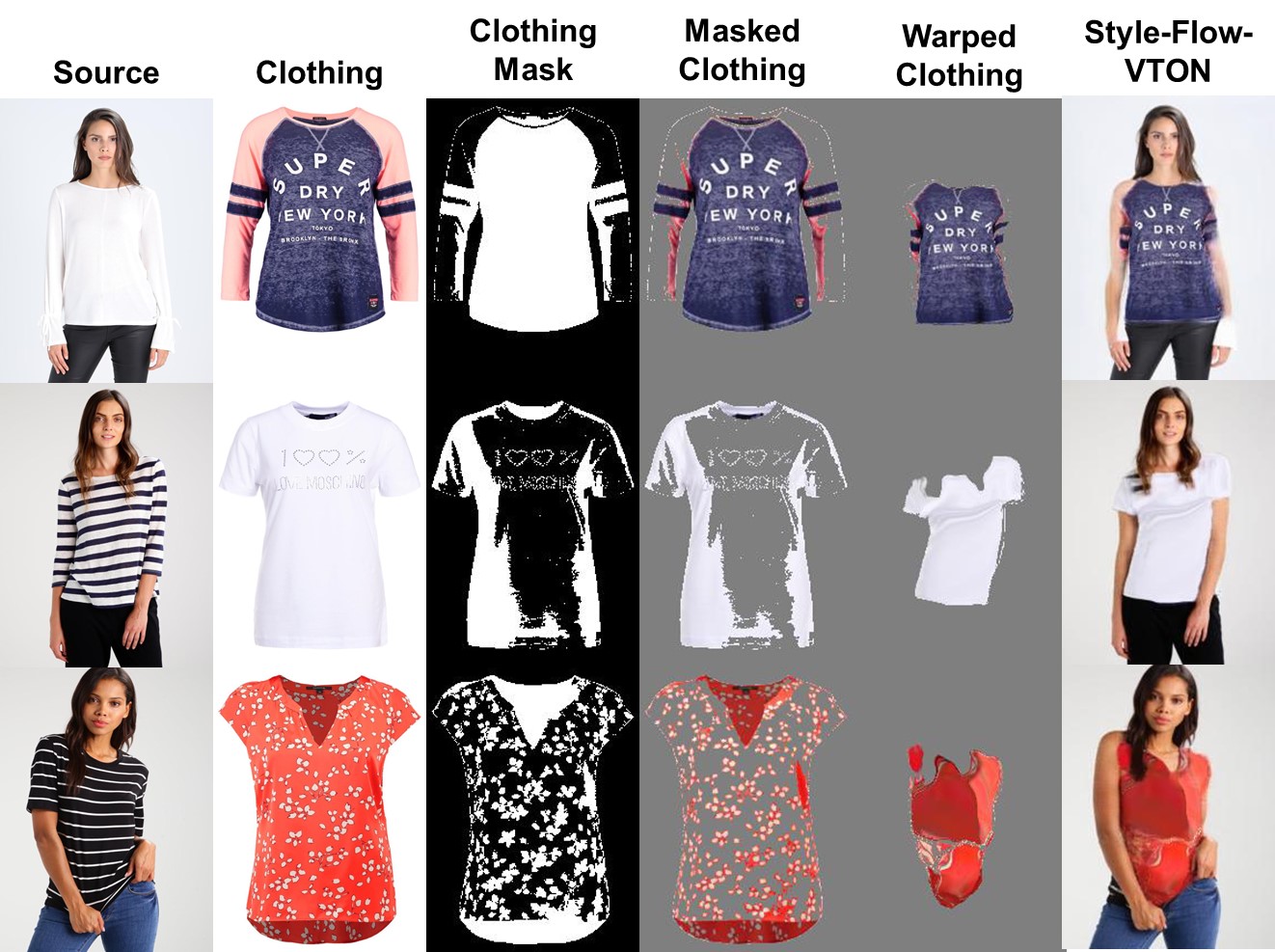}
\caption{Failure cases caused by limited clothing parsing. If the target clothes are masked, it is difficult to restore realistic try-on effects.}
\label{clothmaskcuowu}
\end{figure}

Some wrong clothing masks are shown in Fig. \ref{clothmaskcuowu}. Since Style-Flow-VTON rely on the clothing mask  \citep{in-shop-cloth-mask} to extract ``pure clothing image" without background, it heavily relies on the performance of clothing parsing. For clothing with complex textures or similar color as the background, parsed mask usually contains artifacts, which loses important clothing information for final try-on image.

It is worth mention that clothing parsing  is not necessary, e.g., for recenet diffusion-based models, where only the character image and clothing image are needed as conditions, and a mask is used to guide the area to be tried on. Due to the strong feature parsing ability and scalability of large image generation models, the dependency on parsing will be also alleviated.

\section{Future Work}\label{sec:future}
Current visual results show that there is still significant room for improvement in image-based virtual try-on methods. On one hand, person representation does not totally get rid of the original clothes, which will affect the clothing style of target clothes. On the other hand, clothing warping approaches are not flexible enough, especially for challenging poses. The emergence of diffusion models alleviates this issue to some extent, but controllable generation becomes a new challenge. \reviseMark{In this section, we first continue to show some potential improvements along current research directions in aspects of person representation, clothing deformation, controllable generation, acceleration, datasets and evaluation metrics. Once the quality of generation becomes satisfactory, additional ideas such as mask-free in-the-wild try-on and scenario-specific generation can also be taken into account.}

\noindent\textbf{Clothing-agnostic person representation.}
As the training data is paired where the person already wears target clothes, it will harm the generalization ability if person representation contains clothes information. Current representations $\mathcal{P}_{\substack{3,6,7,10,11}}$ in Fig. \ref{fig:person_representation} contains clues from the original clothes such as clothing contour, which will affect the clothing style of target clothes. Totally deleting the body area (i.e., $\mathcal{P}_{\substack{4,12}}$ will get rid of original clothes but lose person characteristics such as skin color. Constructing triplets where a person wears different clothes could be a solution, but current methods  \citep{WUTON,PFAFN,style-flow} still rely on parser-based methods to construct the triplet, which inherits the limitation of human parser. \reviseMark{Unpaired try-on has been explored, but is still challenging facing the lack of supervised training.} Future direction could try to construct training triplets with parser-free approaches such as recent diffusion models, or totally delete original clothes but keep a sampler at exposed skin. For diffusion models based on mask inpainting, how to decouple the mask with original clothes also deserves exploring.

\noindent\textbf{Natural clothing deformation.} 
Transformation under intrinsic rules such as STN and TPS is limited for flexible deformation. For example, they cannot put sleeves before the torso. Patch-based methods alleviate the limitation to some extent. Flow estimation learns pixel offset which improves flexibility but cannot generates new pixels that are not displayed in target clothing image but are shown under current posture. Implicit transformation shows potentials in clothing deformation and deserves further exploration, especially driven by a 3D deformation model. Additionally, existing methods distort target clothing image into the try-on area and ignore the original style. For example, a cropped top is stretched or a long coat is shrunken to normal length. Fine-grained clothing landmarks could be helpful for this situation.

\noindent\textbf{Controllable generation with diffusion model.}
Diffusion models show powerful generalization and high quality, which is appealing for virtual try-on tasks. However, current models only meet semantic-level control such as satisfying the description from text. More precise and strict control has not been well solved such as preserving the original styles (e.g., Fig. \ref{chuanyipianjian2}). 
Effectively encoding clothing image and injecting to the generation model, or just using diffusion models to refine the warped clothes deserve a try. 

\noindent\textbf{Automation and acceleration.}
Current methods rely on a series of pre-processing steps, which affects the automation and inference time. More parser-free models can be explored in the future. Additionally, with the diffusion model as the dominant generative framework, addressing its time-consuming sampling process deserves attentions.

\noindent\textbf{Multi-modal foundation models.} Besides training a specific large-scale model, how to make existing powerful models, e.g., multi-modal foundation models, facilitate virtual try-on task deserves exploration. By incorporating additional modalities, such as textual descriptions or style guidance, multi-modal foundation models can to some extent improve the controllability and customization of the generated virtual try-on images. 
Furthermore, multi-modal models can leverage the synergistic information from different modalities to enhance the quality of generated images and provide a more immersive and interactive virtual try-on experience.

\noindent\textbf{Diverse datasets.}
Constructing large-scale and diverse virtual try-on datasets forms the foundation of research in this domain. Future datasets should encompass various types and styles of clothing, as well as diverse human body shapes, poses, and skin tones. Particularly, associated descriptions such as clothing style, size and fabric property are always contained in the website but ignored by existing methods. With the development of cross-modality models, cooperation with these descriptions in image generation might bring benefits.

\noindent\textbf{Specialized evaluation metrics.}
Existing evaluation metrics are not specialized for virtual try-on tasks \citep{evaluation_metrics}. In this paper, a new criteria is proposed with advanced pre-trained model and specifically evaluate the semantic information for try-on and non-try-on areas. Future attention could be paid to assessing the clarity and realism of clothing patterns, the preservation of clothing styles and the person characteristics. 

\reviseMark{\noindent\textbf{Mask-free in-the-wild try-on.}
Current in-the-wild try-on methods rely on accurate mask for inpainting and only allows slight occlusions. However, fixed masking strategy is not flexible enough and mask inpainting has the following drawbacks for image-based virtual try-on: 1) It is difficult to preserve the original characteristics such as the skin tone, tattoo or other decorations. 2) The original clothing area that is not masked will cause artifacts. 3) The target clothing is distorted into the inpainting area with the loss of clothing types. Therefore, how to accurately learn the clothing area that is expected to be altered is important for mask-free in-the-wild try-on. It also deserves a try to explore flexible loss functions to support unpaired training, which respectively constrains try-on area with target clothing image and non-try-on area from person image.}

\noindent\textbf{Scenario-specific try-on.}
Real-world try-on is limited to dressing rooms, while try-on effects differ in different scenarios such as indoor or outdoor, different seasons and different occasions. Virtual try-on has advantages in synthesizing the background, which provides imaginary space for users to consider the suitability of target clothes. Furthermore, generating videos to showcase the effects in different scenarios would be more engaging \citep{Action2video}.

\section{Conclusion}
With the rapid development of image generation models, the performance of image-based virtual try-on task is getting closer to practical applications. To the best of our knowledge, this is the first systematic review for image-based virtual try-on with unified evaluation and in-depth analysis. In this review, we first propose taxonomy in aspects of try-on indication, cloth warping and try-on modules and give in-depth analysis towards representative methods in these designs. Afterwards, we introduce datasets and evaluation criteria, and uniformly evaluate representative methods using mainstream and new proposed criteria. Quantitative, qualitative and user study experiments are conducted and analyzed. Finally, we show current unresolved issues and point out future directions. We hope this survey provides an effective way to comprehensively understand image-based virtual try-on and inspires further exploration of this research field. 

\bmhead{Acknowledgements}
This work was supported in part by the National Natural Science Foundation of China under Grant U21B2024 and Grant 61902277.


\section*{Declarations}
\begin{itemize}
    \item \textbf{Competing interests.} The authors have no competing interests to declare that are relevant to the content of this article.
    \item \textbf{Data availability.}  The uniformly implemented evaluation metrics, dataset and collected methods will be made public available at https://github.com/little-misfit/Survey-Of-Virtual-Try-On.
\end{itemize}

\bibliography{sn-bibliography}

\end{document}